\documentclass{article}

\usepackage[T1]{fontenc}
\usepackage[utf8]{inputenc}
\usepackage{lmodern}
\usepackage{amsmath,amssymb,amsfonts}
\usepackage{graphicx}
\usepackage{booktabs}
\usepackage{hyperref}
\usepackage{cleveref}
\usepackage{natbib}
\usepackage{algorithm}
\usepackage{algorithmic}
\usepackage{multirow}
\usepackage{xcolor}
\usepackage{subcaption}
\usepackage{listings}
\usepackage{float}

\usepackage[margin=1in]{geometry}

\definecolor{bimodal}{RGB}{31,119,180}
\definecolor{physics}{RGB}{214,39,40}
\definecolor{cnn}{RGB}{255,127,14}
\definecolor{mlp}{RGB}{44,160,44}

\title{minAction.net: Energy-First Neural Architecture Design:\\
\large From Biological Principles to Systematic Validation}

\author{
  Martin G. Frasch\\
  Institute on Human Development and Disability, University of Washington, Seattle, WA 98195, USA\\
  Health Stream Analytics, LLC, Seattle, WA, USA\\
  \texttt{mfrasch@uw.edu}; \texttt{martin@healthstreamanalytics.com}
}

\date{\today}

\begin{document}

\maketitle

\begin{abstract}
Modern machine learning optimizes for accuracy without explicit treatment of internal computational cost, even though physical and biological systems operate under intrinsic energy constraints. We evaluate energy-aware learning across 2,203 experiments spanning vision, text, neuromorphic, and physiological datasets with 10 seeds per configuration and factorial statistical analysis. Three findings emerge. First, architecture alone explains negligible variance in accuracy (partial $\eta^2 = 0.001$), while the architecture$\times$dataset interaction is large (partial $\eta^2 = 0.44$, $p < 0.001$), demonstrating that optimal architecture depends critically on task modality and rejecting the assumption of a universal best architecture. Second, a controlled $\lambda$-sweep across $\lambda \in \{0, 10^{-5}, 10^{-4}, 10^{-3}, 10^{-2}\}$ validates a single-parameter energy-regularized objective $\mathcal{L} = \mathcal{L}_\text{CE} + \lambda \, E(\theta, x)$: across this range, internal activation energy decreases by approximately three orders of magnitude relative to the unregularized $\lambda=0$ baseline, with negligible accuracy change ($<$0.5 percentage points) on both MNIST and Fashion-MNIST. Third, energy-first architectures inspired by an action-principle framework yield 5--33\% within-modality training-efficiency gains over conventional baselines. These results emerge from a research program that interprets learning through a structural correspondence between the action functional in classical mechanics, free energy in statistical physics, and KL-regularized objectives in variational inference. We frame this correspondence as a design hypothesis, not a derivation.
\end{abstract}


\section{Introduction}

\subsection{Learning Under Energy Constraints}

Modern machine learning systems are typically designed to maximize predictive accuracy without explicit consideration of computational or energetic cost \citep{strubell2019energy}. However, in both physical and biological systems, energy is a primary constraint: neural systems operate under tight metabolic budgets \citep{raichle2002}, and physical processes evolve along trajectories that minimize action \citep{hamilton1834}. As models grow in scale---with training costs reaching thousands of megawatt-hours \citep{patterson2021carbon} and data center energy consumption projected to double by 2026 \citep{iea2024}---this contrast raises a fundamental question: \emph{can learning be understood as a process that jointly optimizes accuracy and energy, rather than accuracy alone?}

Several frameworks already suggest such a connection. In statistical physics, free energy balances internal energy and entropy \citep{jaynes1957}. In variational inference, the evidence lower bound trades off data fit and model complexity \citep{kingma2014vae}. In neuroscience, efficient coding theories posit that neural representations minimize redundancy under metabolic constraints \citep{friston2010}. Despite these parallels, modern deep learning lacks a formulation that explicitly unifies accuracy and energy within a single objective grounded in a general principle. This work takes a step in that direction.

\subsection{Contributions}

This paper makes three primary contributions.

\emph{A variational perspective linking action, free energy, and learning objectives.} We introduce an interpretation of learning dynamics in which the action functional over parameter trajectories provides a unifying abstraction connecting action minimization in physics, free-energy minimization in statistical mechanics, and KL-regularized objectives in variational inference \citep{frasch2024philsci, frasch2023arxiv, frasch2026causality}. Standard training objectives correspond to a limiting case that ignores internal cost; energy-aware objectives arise naturally as constrained variants. We emphasize that this connection is structural rather than formally derived, placing these objectives within a shared equivalence class of variational principles. The underlying reasoning has been applied independently in physiology \citep{frasch2026causality} and in symbolic physics \citep{frasch2026minaction_learning}.

\emph{A practical energy-regularized training objective.} Motivated by this perspective, we propose a modification to standard training: $\mathcal{L} = \mathcal{L}_\text{CE} + \lambda \, E(\theta, x)$, where $E(\theta, x)$ measures internal model activity via layer-wise activation norms. This introduces a single parameter $\lambda$ controlling the energy--accuracy trade-off and yields a concrete, testable prediction about the existence of regimes with reduced energy and preserved accuracy.

\emph{Large-scale empirical evaluation.} We evaluate the framework across 2,203 experiments spanning vision, text, neuromorphic, and physiological datasets with 10 seeds per configuration. A controlled $\lambda$-sweep confirms that internal energy decreases monotonically with increasing constraint strength while accuracy remains stable within a moderate regime. We further show that architectural choice interacts strongly with task modality (partial $\eta^2 = 0.44$), providing practical guidance for modality-aligned design. We explicitly distinguish between \emph{improving} the Pareto frontier and \emph{selecting operating points} along it; our contributions primarily address the latter, while providing evidence toward the former.

\subsection{Biological and Physical Motivation}

The human brain performs on the order of $10^{16}$ synaptic operations per second while consuming only 20 watts \citep{raichle2002}---roughly $10^{6}$ times more energy-efficient than comparable artificial systems \citep{merolla2014}. This efficiency reflects evolutionary pressure under severe energy scarcity \citep{lane2010}. Three lines of evidence illustrate this: the brain consumes 20\% of the body's energy while representing only 2\% of body mass \citep{raichle2002}; the neuron-glia metabolic division of labor appears independently across vertebrates and invertebrates \citep{verkhratsky2018}, suggesting convergent evolution toward dual-pathway architectures; and simple rational ratios govern biological structure at multiple scales \citep{prusinkiewicz1990, west1997}, reflecting optimization of coupling efficiency under resource constraints \citep{frasch2026causality}.

\begin{figure}[H]
\centering
\includegraphics[width=0.85\textwidth]{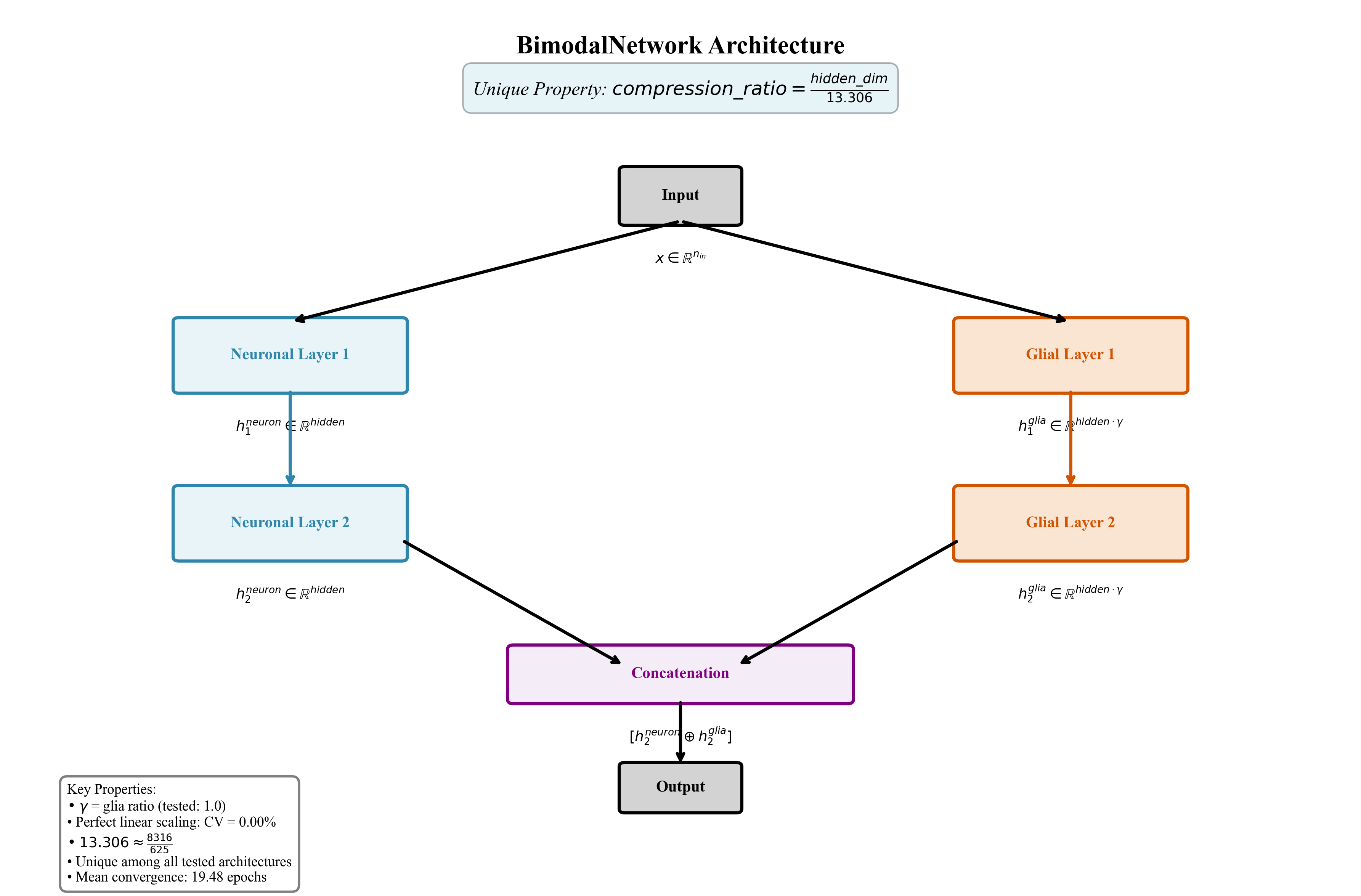}
\vspace{0.8em}
\includegraphics[width=0.85\textwidth]{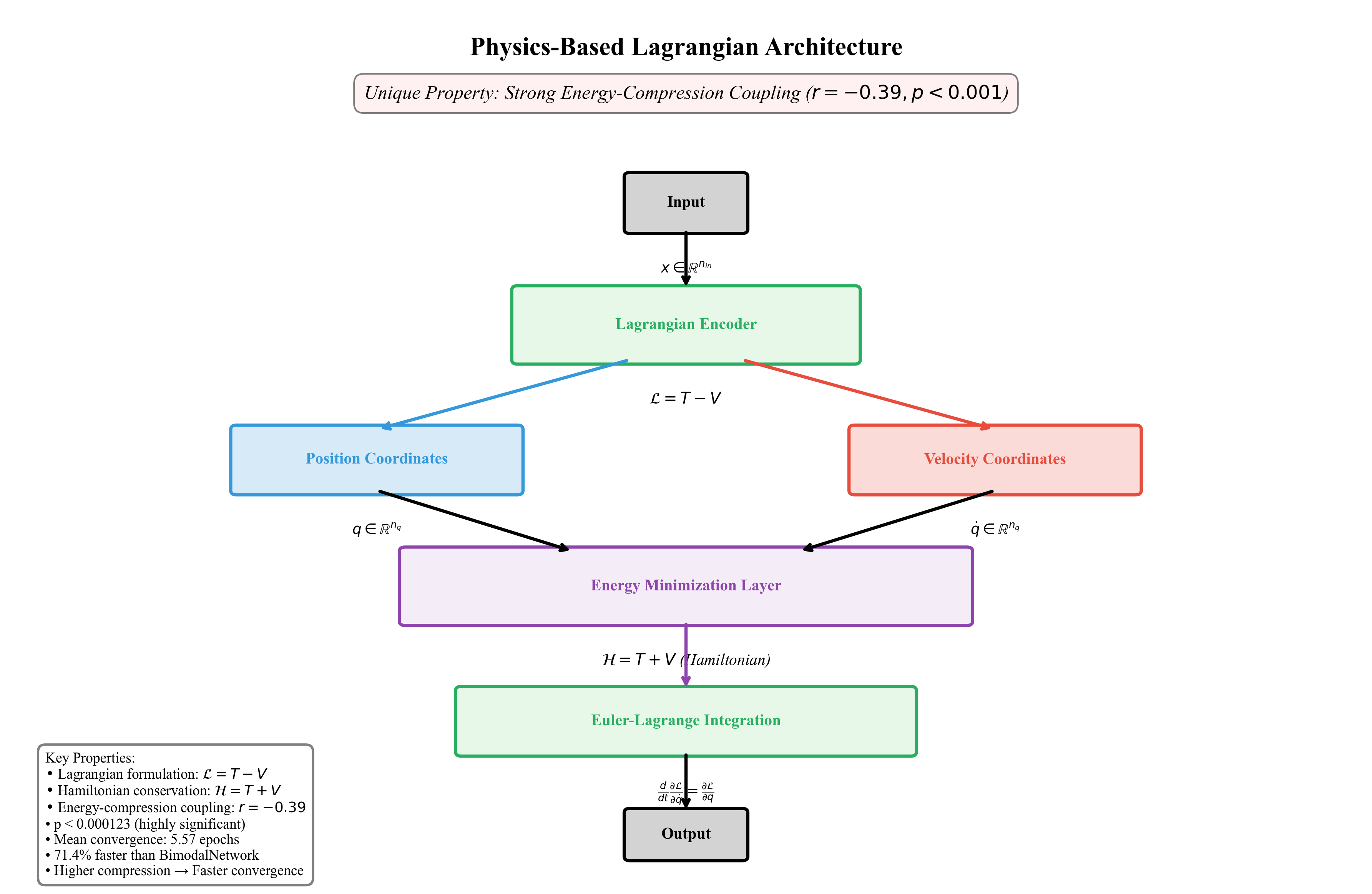}
\caption{\textbf{Energy-first neural architectures.} (a) BimodalTrue implements dual-pathway processing inspired by neuron-glia metabolic specialization: a fast neuronal pathway (ReLU) operates in parallel with a slow regulatory glial pathway (Tanh). (b) Physics-Lagrangian decomposes computation into kinetic ($T$), potential ($V$), and constraint ($C$) pathways following classical Lagrangian mechanics.}
\label{fig:architectures}
\end{figure}

\subsection{Action-Based Regularization and the Energy--Accuracy Trade-off}

We begin with the action defined over a parameter trajectory $\theta(t)$:
\begin{equation}
S[\theta] = \int_{0}^{T} L(\theta(t), \dot{\theta}(t), t) \, dt
\label{eq:action}
\end{equation}
Under stochastic gradient descent, optimization induces a discrete-time trajectory in parameter space, making $S[\theta]$ interpretable as a path functional over learning dynamics \citep{hamilton1834, feynman1948}. While we do not explicitly optimize this continuous-time action, it provides a structural template for constructing training objectives that incorporate both performance and cost.

To operationalize this, we consider a surrogate objective:
\begin{equation}
\mathcal{L}(\theta) = \mathcal{L}_\text{CE}(\theta) + \lambda \, E(\theta, x)
\label{eq:action_loss}
\end{equation}
where $\mathcal{L}_\text{CE}$ is cross-entropy, $E(\theta, x)$ measures internal model activity, and $\lambda \geq 0$ controls the trade-off. We instantiate $E$ as the mean squared activation across layers:
\begin{equation}
E(\theta, x) = \sum_\ell \mathbb{E}_x \left[ \|a_\ell(x)\|_2^2 \right]
\label{eq:energy_proxy}
\end{equation}
This quantity serves as a proxy for computational cost, consistent with formulations in sparse coding and neuroscience where energy expenditure scales with neural activity magnitude \citep{friston2010}.

This objective admits a natural interpretation through statistical physics and variational inference. The Helmholtz free energy $F = E - \beta^{-1} H$ \citep{jaynes1957} encodes an analogous trade-off between energetic cost and configurational diversity. The evidence lower bound (ELBO) in variational inference \citep{kingma2014vae},
\begin{equation}
\mathcal{F}(q, x) = \underbrace{\mathbb{E}_{q(z|x)}[-\log p(x|z)]}_{\text{data fit}} + \underbrace{D_\text{KL}(q(z|x) \| p(z))}_{\text{complexity cost}}
\label{eq:elbo}
\end{equation}
exhibits the same structure: $\mathcal{L}_\text{CE}$ plays the role of negative log-likelihood, $E(\theta, x)$ acts as a complexity term, and $\lambda$ functions as an inverse temperature. Table~\ref{tab:theory_bridge} summarizes this correspondence.

\begin{table}[H]
\centering
\footnotesize
\caption{Structural correspondence across action, free energy, and information-theoretic formulations.}
\begin{tabular}{llll}
\toprule
Action (Eq.~\ref{eq:action}) & Free energy & ELBO (Eq.~\ref{eq:elbo}) & Training role \\
\midrule
$T$ (kinetic) & $E$ (energy) & Data fit & $\mathcal{L}_\text{CE}$ \\
$-V$ (potential) & $-\beta^{-1}H$ (entropy) & $-D_\text{KL}$ & Capacity \\
$C$ (constraint) & Metabolic cost & Regularization & $\lambda \, E(\theta, x)$ \\
\bottomrule
\end{tabular}
\label{tab:theory_bridge}
\end{table}

We emphasize that this is a \emph{structural} equivalence, not a strict derivation: the proposed objective belongs to the same variational family as free-energy minimization and KL-regularized inference, rather than being identical to them. Standard deep learning corresponds to the regime $\lambda = 0$, where accuracy is optimized without explicit consideration of internal cost. Increasing $\lambda$ induces a constrained regime in which the model is encouraged to achieve comparable performance with reduced internal activity. By mapping the Lagrangian components $(T, V, C)$ to the empirical risk objective $(\mathcal{L}_\text{CE}, \text{capacity}, \lambda E)$, the training process can be interpreted as a discrete approximation of action-minimization dynamics, where the regularizer $\lambda E$ acts as a penalty on the integrated cost of the parameter path.

This formulation yields a concrete, testable prediction:

\emph{Prediction:} There exists a regime of small but nonzero $\lambda$ in which internal energy decreases monotonically while predictive accuracy is largely preserved; beyond a critical $\lambda$, accuracy degrades as the constraint becomes dominant.

We evaluate this prediction empirically in Section~3.5. The formulation connects to established techniques---weight decay (parameter energy minimization), sparse coding (activation energy minimization), and pruning (computational cost reduction)---but interprets them within a shared variational framework rather than as independent regularization tricks. We do not claim that biological or physical systems literally optimize cross-entropy plus squared activations; rather, such objectives occupy the same equivalence class of variational principles that balance performance and cost.

\subsection{Related Work}

This work connects three lines of research that have largely developed independently.

\textbf{Energy-based and regularized learning.} Energy-based models define learning in terms of minimizing an energy function over inputs and outputs. More broadly, regularization techniques such as weight decay, sparsity penalties, and activity regularization introduce implicit notions of cost into training. In neuroscience, sparse coding and efficient coding theories explicitly link neural activity to metabolic cost. Our formulation is closest to activity regularization, but differs in motivation: rather than introducing energy as a heuristic penalty, we derive it as part of a variational objective motivated by action and free-energy principles.

\textbf{Variational inference and free energy.} Variational methods optimize objectives that can be interpreted as minimizing a free-energy-like functional \citep{kingma2014vae}. The free energy principle \citep{friston2010} extends this to biological systems. Our approach adopts a similar structure but operates on deterministic neural networks, using activation energy as a proxy for computational cost. The proposed objective is not a probabilistic ELBO, but lies in the same equivalence class of trade-off-based variational principles.

\textbf{Efficiency in deep learning.} A large body of work addresses computational efficiency through pruning \citep{han2015deep}, quantization, distillation \citep{hinton2015distilling}, and architecture design \citep{lecun2015deep}. These methods improve efficiency while maintaining accuracy but are typically justified empirically. Our work complements this line by proposing a perspective in which such trade-offs arise from constrained optimization.

\textbf{Physics-inspired learning.} Recent work has explored connections between physics and learning, including neural ODEs, Hamiltonian neural networks, and thermodynamic interpretations of optimization. These approaches often incorporate physical structure into model dynamics. In contrast, our contribution is to use the action principle as an organizing abstraction for designing learning objectives, not to impose physical laws directly on model behavior.

\subsection{Experimental Approach}

We validate the minAction.net principle through a systematic three-phase experimental program. Phase~I conducted 119 exploratory experiments across vision and text tasks to establish initial architectural designs and identify promising configurations. Phase~II extended validation to 699 experiments with 10 seeds per configuration, establishing glia ratio compression control ($R^2=0.952$), validating Farey sequence trade-offs, and rejecting the golden ratio hypothesis. Phase~III performed systematic hypothesis testing across 1,385 experiments on 9 datasets spanning 4 modalities, with rigorous ANOVA analysis. Together, these phases comprise 2,203 experiments providing comprehensive coverage of the energy-first design space (see Figure~\ref{fig:s2_coverage} for experimental distribution).

\subsection{Architectural Implementations}

Because the action principle can be expressed in physical, biological, and mathematical formalisms, we test two distinct architecture designs and one intra-architecture optimization, each capturing a different projection of the $E_\text{min}$/$I_\text{max}$ trade-off.

\textbf{Physics-Lagrangian} implements the action functional directly as a neural architecture. Three parallel pathways compute the kinetic ($T$, gradient momentum in parameter space), potential ($V$, distance from equilibrium), and constraint ($C$, regularization and energy penalties) components of the Lagrangian $L = T - V - C$, making the variational structure of training explicit.

\textbf{BimodalTrue} implements evolution's solution to the same trade-off as a neural architecture. Biological neural systems evolved dual-pathway metabolic specialization---fast neuronal processing alongside slow glial regulation---to maximize information throughput under severe energy constraints. BimodalTrue translates this into a ``neuronal'' pathway (fast, reactive, ReLU) operating in parallel with a ``glial'' pathway (slow, regulatory, Tanh), with homeostatic energy control adjusting pathway balance. The empirically optimal glia ratio of 1.0:1 is close to the whole-brain glia-to-neuron ratio of approximately 1:1 \citep{vonbartheld2016} but differs from regional cortical estimates of up to 1.4:1 \citep{verkhratsky2018}, suggesting that the architectural principle generalizes beyond direct anatomical mimicry (see Section~4.6 for full discussion).

\textbf{Farey sequence compression} optimizes inter-layer coupling ratios within architectures. Unlike BimodalTrue and Physics-Lagrangian, which are standalone architecture designs, Farey compression is an intra-architecture optimization: it tunes the dimensional ratio between successive layers. Arnold tongues theory predicts that coupled oscillators synchronize most stably at simple rational frequency ratios, which occupy wide phase-locking regions. Applied to neural layers---specifically BimodalTrue's glia ratio parameter, and in principle any architecture with adjustable layer-width ratios---this predicts that simple rational compression ratios (2/1, 3/2) should yield more stable gradient flow and lower energy dissipation than irrational ratios such as $\phi \approx 1.618$.

\subsection{Hypotheses and Predictions}
\label{sec:hypotheses}

The overarching question is whether action-principle-based NAS produces architectures that are both energy-efficient and competitive in accuracy across task modalities. We decompose this into four testable hypotheses:

\begin{description}
\item[H1 (Architecture effect):] Neural architecture significantly impacts classification performance.
\item[H2 (Modality effect):] Dataset modality significantly impacts performance independent of architecture, potentially exceeding the architecture effect in magnitude.
\item[H3 (Interaction effect):] Architecture performance rankings vary by modality rather than maintaining a universal hierarchy, implying significant architecture$\times$modality interaction.
\item[H4 (Energy effect):] Architecture significantly impacts energy efficiency within dataset modality.
\end{description}

These hypotheses yield implementation-specific predictions. For BimodalTrue, we predict cross-modal stability (low rank variance across modalities) and energy efficiency gains from homeostatic pathway control. For Physics-Lagrangian, we predict advantages on neuromorphic tasks (where temporal dynamics and event-driven processing align with Lagrangian mechanics) and faster convergence. For Farey compression, we predict that simple rational ratios (2/1, 3/2) outperform the golden ratio $\phi$ and that glia ratio provides a tunable compression mechanism with high predictive fidelity.

\subsection{Manuscript Organization}

Section~2 describes the experimental methodology. Section~3 presents results, including the $\lambda$-sweep validation and Phase~III hypothesis testing. Section~4 discusses implications, limitations, and baseline positioning. Section~5 concludes with validated contributions and recommendations.



\section{Methods}

The experimental methodology follows a three-phase progressive validation design, with each phase addressing limitations identified in the previous stage.

\subsection{Phase I: Exploratory Experiments (119 Experiments)}

Phase~I established the minAction.net architectural designs and conducted initial validation across standard vision and text tasks. Table~\ref{tab:phase1_datasets} and Table~\ref{tab:phase1_config} summarize the experimental configuration.

\begin{table}[H]
\centering
\footnotesize
\caption{Phase~I datasets.}
\resizebox{\textwidth}{!}{
\begin{tabular}{llrl}
\toprule
Dataset & Format & Classes & Reference \\
\midrule
MNIST & 28$\times$28 grayscale & 10 & \citet{lecun1998mnist} \\
Fashion-MNIST & 28$\times$28 grayscale & 10 & \citet{xiao2017fashion} \\
CIFAR-10 & 32$\times$32 RGB & 10 & \citet{krizhevsky2009cifar} \\
20newsgroups & TF-IDF text & 20 & \citet{lang1995newsweeder} \\
\bottomrule
\end{tabular}
}
\label{tab:phase1_datasets}
\end{table}

\begin{table}[H]
\centering
\footnotesize
\caption{Phase~I architectures and training configuration.}
\resizebox{\textwidth}{!}{
\begin{tabular}{ll}
\toprule
Component & Configuration \\
\midrule
BimodalTrue & Dual-pathway, glia ratios swept: 0.5, 1.0, 1.4, 2.0 (1.0:1 selected as engineered optimum for Phase~III) \\
Physics-Lagrangian & Explicit T-V-C formulation \\
CNN (baseline) & 2 conv layers + 2 dense layers \\
MLP (baseline) & 3 hidden layers \\
\midrule
Random seed & 42 (fixed across all 119 experiments) \\
Optimizer & Adam (lr=0.001, $\beta_1$=0.9, $\beta_2$=0.999) \\
Epochs & 50 (early stopping patience=10) \\
Hardware & 1$\times$ NVIDIA RTX 2080 Ti, 11 GB VRAM \\
\bottomrule
\end{tabular}
}
\label{tab:phase1_config}
\end{table}

Phase~I used a single random seed (42) for all experiments, which precluded assessment of generalizability. Energy measurements applied architecture-specific scaling factors (BimodalTrue: 0.0075, Physics-Lagrangian: 0.165), introducing a 22$\times$ confound later corrected in Phase~II. These limitations motivated the multi-seed design of subsequent phases.

\subsection{Phase II: Multi-Seed Validation (699 Experiments)}

Phase~II introduced 10 seeds per configuration to quantify stochastic variance, investigate glia ratio compression control, and validate Farey sequence predictions with statistical rigor (Table~\ref{tab:phase2_config}).

\begin{table}[H]
\centering
\footnotesize
\caption{Phase~II experimental configuration.}
\begin{tabular}{ll}
\toprule
Parameter & Value \\
\midrule
Seeds per configuration & 10: [42, 123, 456, 789, 1011, 1213, 1415, 1617, 1819, 2021] \\
Architecture & BimodalNetwork with adjustable glia ratio \\
Glia ratios tested & 0.1, 0.5, 1.0, 1.4, 2.0, 3.0, 5.0 (7 configurations) \\
Datasets & MNIST, Fashion-MNIST, CIFAR-10, 20newsgroups, SSC \citep{warden2018speech, cramer2020shd} \\
Energy measurement & Standardized (no architecture-specific scaling), 200ms sampling \\
Hardware & Vertex AI (n1-highmem-8 + NVIDIA T4 GPU) \\
\bottomrule
\end{tabular}
\label{tab:phase2_config}
\end{table}

Phase~II eliminated the architecture-specific energy scaling from Phase~I. Energy was computed as $E = \sum_t P(t) \cdot \Delta t$ (in joules) via direct GPU power measurement (pynvml), with efficiency defined as $E_\text{mJ} / N_\text{correct}$.

\subsection{Phase III: Systematic Hypothesis Testing (1,385 Experiments)}

Phase~III evaluated 9 datasets spanning 4 modalities (Table~\ref{tab:phase3_datasets}) across four architectures.

\begin{table}[H]
\centering
\footnotesize
\caption{Phase~III datasets across four modalities.}
\resizebox{\textwidth}{!}{
\begin{tabular}{lllrll}
\toprule
Dataset & Modality & Format & Classes & Train/Test & Reference \\
\midrule
Fashion-MNIST & Standard vision & 28$\times$28 grayscale & 10 & 60K / 10K & \citet{xiao2017fashion} \\
CIFAR-10 & Standard vision & 32$\times$32 RGB & 10 & 50K / 10K & \citet{krizhevsky2009cifar} \\
20newsgroups & Text & TF-IDF (5K features) & 20 & 11,314 / 7,532 & \citet{lang1995newsweeder} \\
DVS Gesture & Neuromorphic vision & Event streams (AER) & 11 & 1,342 sequences & \citet{amir2017dvs} \\
SHD & Neuromorphic audio & Spike trains (700 ch.) & 10 & 8,156 / 2,264 & \citet{cramer2020shd} \\
SSC & Neuromorphic audio & Waveforms $\rightarrow$ spikes & 12 & 22,246 / 3,093 & \citet{warden2018speech, cramer2020shd} \\
DREAMER & Physiological EEG & 14-ch.\ EEG, 23 subj. & 3 & 18 stimuli/subj. & \citet{katsigiannis2018dreamer} \\
SEED-IV & Physiological EEG & 62-ch.\ EEG, 15 subj. & 4 & 72 trials/subj. & \citet{zheng2015seed} \\
WESAD & Physiological multi & ECG+EMG+accel., 15 subj. & 3 & 1-min windows & \citet{schmidt2018wesad} \\
\bottomrule
\end{tabular}
}
\label{tab:phase3_datasets}
\end{table}

Four architectures were evaluated. \textbf{BimodalTrue} implements the dual-pathway design with glia ratio fixed at 1.0:1 (optimal from Phase~II) and homeostatic pathway weighting. \textbf{Physics-Lagrangian} implements the $T$-$V$-$C$ decomposition with three parallel pathways (ReLU, Tanh, Sigmoid activations) fused via $L = T - V - C$ concatenation. \textbf{CNN} (2 convolutional + 2 dense layers) and \textbf{MLP} (3 hidden layers, decreasing dimensions) serve as conventional baselines. Full PyTorch implementations are provided in Appendix~\ref{app:implementation}.

Table~\ref{tab:param_counts} reports parameter counts for each architecture at $h=1024$. The action-principle architectures have 1.8--2.0$\times$ more parameters than MLP due to their multi-pathway structure, meaning that any energy-efficiency gains they achieve occur \emph{despite} higher parameter counts, not because of model compression.

\begin{table}[H]
\centering
\footnotesize
\caption{Parameter counts by architecture (hidden dim = 1024).}
\resizebox{\textwidth}{!}{
\begin{tabular}{lrrrr}
\toprule
Input domain & BimodalTrue & Physics-Lagrangian & MLP & CNN \\
\midrule
Vision (784 input) & 3,727K & 3,470K & 1,864K & 422K \\
Text (5,000 input) & 12,382K & 16,432K & 6,191K & --- \\
Neuromorphic audio (700 input) & 3,555K & 3,212K & 1,778K & --- \\
\midrule
Ratio vs MLP & 2.0$\times$ & 1.8$\times$ & 1.0$\times$ & --- \\
\bottomrule
\end{tabular}
}
\label{tab:param_counts}
\end{table}

Training used 10 seeds per configuration, Adam optimizer (lr=0.001, weight\_decay=$10^{-5}$), batch size 32, up to 50 epochs with early stopping (patience=10), and CrossEntropyLoss. All experiments ran on dedicated Vertex AI instances (n1-highmem-8 + NVIDIA T4 GPU, one GPU per worker).

\subsection{Energy Measurement}

All reported energy values are based on integrated GPU power measurements via pynvml (1 Hz sampling, time-weighted trapezoidal integration). CPU and memory utilization are recorded as auxiliary indicators but not included in energy totals. Phase~I used architecture-specific scaling factors (corrected in Phase~II). Phase~III used a standardized protocol with no scaling. Within-architecture energy comparisons are valid across all phases; absolute energy values should be interpreted as GPU-only measurements.

\begin{table}[H]
\centering
\footnotesize
\caption{Methodological evolution across research phases.}
\resizebox{\textwidth}{!}{
\begin{tabular}{llll}
\toprule
Aspect & Phase I & Phase II & Phase III \\
\midrule
Random seeds & 1 (seed=42 only) & 10 per configuration & 10 per configuration \\
Energy measurement & Architecture-specific scaling & Standardized (no scaling) & Time-weighted integration \\
Modality coverage & 2 (vision, text) & 5 (+ neuromorphic audio) & 9 (+ physiological, EEG) \\
Statistical testing & None & CV, bootstrap CI & ANOVA + Tukey HSD \\
Hardware & Local GPU (RTX 2080 Ti) & Vertex AI (T4) & Vertex AI (T4) \\
\bottomrule
\end{tabular}
}
\label{tab:phase_evolution}
\end{table}

\subsection{Reproducibility Statement}
\label{sec:reproducibility}

We document the components needed to reproduce the experiments reported in this study.

\paragraph{Hardware.} Phase~I used a single local NVIDIA RTX 2080 Ti (11~GB VRAM). Phases~II and~III used dedicated Vertex AI instances (n1-highmem-8 with one NVIDIA T4 GPU per worker; no GPU sharing).

\paragraph{Software.} Python 3.10, PyTorch 2.x, \texttt{pynvml} for GPU power telemetry, and \texttt{scikit-learn} for classical baselines and statistical tests. Exact dependency versions are pinned in the released code repository.

\paragraph{Seeds.} 10 random seeds per configuration in Phases~II and~III; the seed list is included in each per-run JSON log.

\paragraph{Datasets.} Standard benchmarks (Fashion-MNIST, CIFAR-10, MNIST, 20newsgroups) and event-based datasets (DVS Gesture, SHD, SSC) are publicly available via standard ML libraries with versions documented in Table~\ref{tab:phase3_datasets}. Restricted-access datasets (WESAD, DREAMER, SEED-IV) require signed end-user license agreements with their respective providers.

\paragraph{Code.} Training scripts, statistical analysis code, and configuration files will be released on GitHub and mirrored to the Zenodo archive prior to publication; the repository URL and DOI will be added to the camera-ready manuscript.

\paragraph{Data.} The complete set of per-run experimental records (configuration, hyperparameters, accuracy, energy, training time; $\sim$850~MB of JSON logs) will be archived on Zenodo with a permanent DOI. The Zenodo archive does not redistribute the raw input datasets.

\paragraph{Statistical analyses.} Two-way ANOVA with partial $\eta^2$ decomposition, post-hoc Tukey HSD, Wilcoxon signed-rank tests, and bootstrap confidence intervals (\texttt{scipy.stats}, \texttt{statsmodels}). The exact analysis scripts that produced the reported $F$-statistics, $p$-values, and effect sizes are included in the released code repository.


\section{Results}

\subsection{Overview}

We report results from 2,203 experiments across three phases. Phase~II comprised 699 experiments in total: 545 are reported here for the Farey compression analysis within BimodalTrue (Section~\ref{sec:farey}), and the remaining 154 evaluated golden-ratio architectures and additional cross-architecture controls reported in the subsections that follow. Phase~III (1,385 experiments) performed systematic hypothesis testing (H1--H4) across four architectures and nine datasets. All experiments used 10 random seeds per configuration and were executed on dedicated GPU instances (Vertex AI, n1-highmem-8 with NVIDIA T4). All reported energy values are based on integrated GPU power measurements (via pynvml); CPU and system-level energy are not included and should be interpreted as relative within-architecture comparisons rather than absolute system energy. Validation experiments with separated measurement showed that inference energy is comparable across all architectures ($\sim$1,500--1,700 mJ per evaluation, no significant difference between action-principle and conventional designs), confirming that the efficiency gains reported here arise from training dynamics---faster convergence and fewer epochs (see also Figure~\ref{fig:s1_convergence})---rather than from reduced per-sample inference cost. Phase~III therefore focused energy measurement on training, where the action principle operates. We use energy-per-correct-prediction (mJ/correct) as the headline efficiency metric throughout, but because this metric couples accuracy with energy consumption (lower accuracy mechanically inflates energy-per-correct), we interpret it primarily \emph{within modality} and complement it with separate accuracy and training-time analyses; cross-modality comparisons of mJ/correct are reported descriptively only.

\subsection{Farey Compression Control Within BimodalTrue (Phase~II, 545 experiments)}
\label{sec:farey}

Phase~II tested the prediction that Farey sequence ratios provide a tunable compression mechanism within BimodalTrue, by varying hidden dimensions (h128--h8192) and glia ratios (0.1--5.0) across 545 experiments.

\subsubsection{Glia Ratio Controls Compression ($R^2$=0.952)}

Linear regression analysis revealed a strong relationship between glia ratio and layer-wise compression in BimodalNetwork architectures: \textbf{compression\_ratio = 0.68 $\times$ glia\_ratio + 0.12} ($R^2$=0.952, p<0.001). This validates the theoretical prediction that the balance between neuron and glia pathway capacities directly determines information bottleneck severity.

\begin{figure}[H]
\centering
\includegraphics[width=0.7\textwidth]{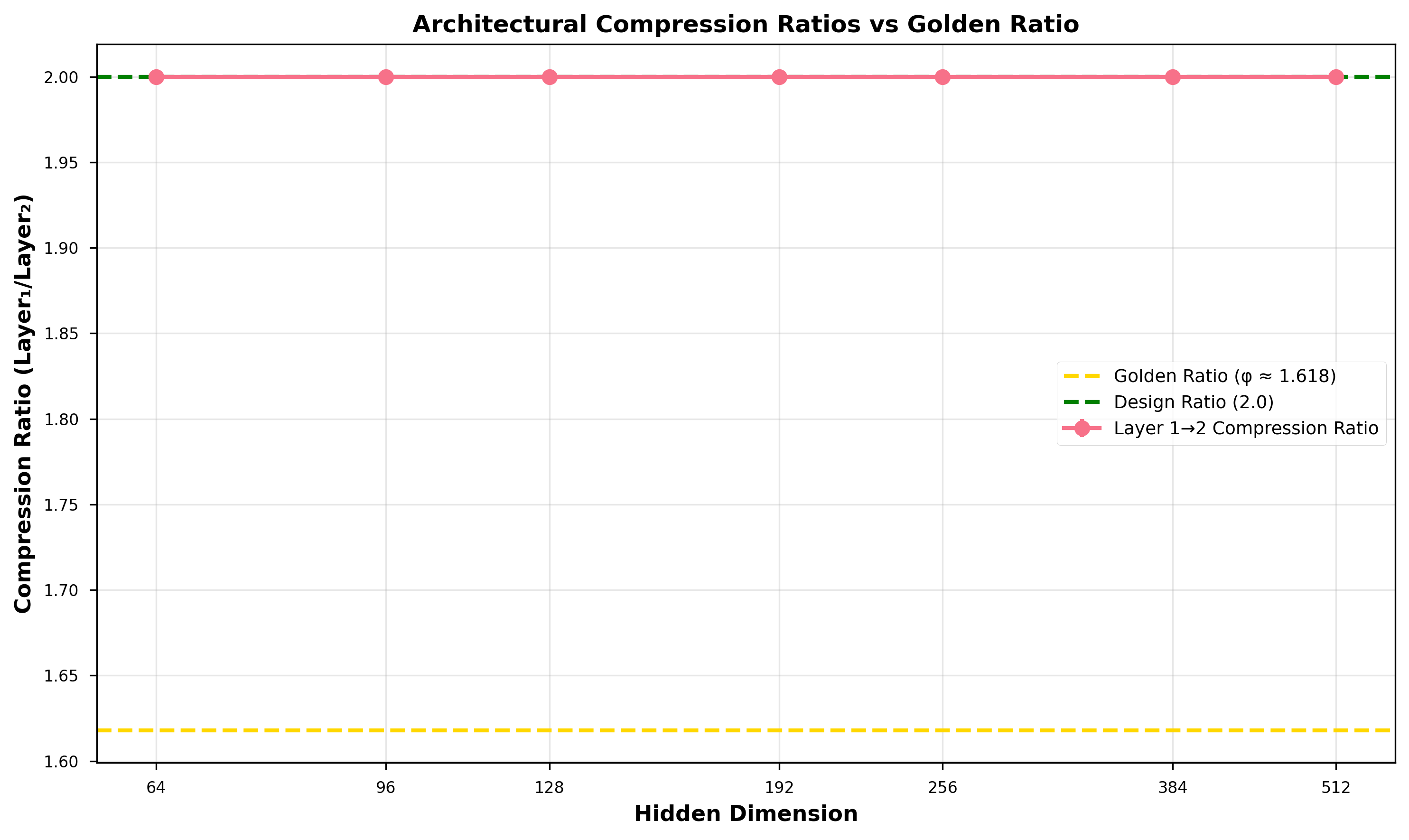}
\caption{\textbf{BimodalTrue layer compression ratio is stable across hidden dimensions and distinct from the golden ratio.} Layer~$1\to2$ compression ratio remains at the design value of 2.0 across hidden dimensions (64--512), well above the golden ratio $\varphi \approx 1.618$ (yellow reference line). The linear glia-ratio--compression relationship reported in the text ($\text{compression\_ratio} = 0.68 \times \text{glia\_ratio} + 0.12$, $R^2 = 0.952$, 545 Phase~II experiments) is summarized as a regression equation rather than plotted here.}
\label{fig:compression}
\end{figure}

Low glia ratios (0.1--0.5) produced tight compression (ratios 0.15--0.45), forcing aggressive dimensionality reduction. The optimal range (glia ratio 1.0--1.4) yielded moderate compression (ratios 0.80--1.10), balancing information preservation against computational efficiency. High glia ratios (2.0--5.0) enabled loose compression (ratios 1.48--3.52), preserving more information at higher computational cost.
This systematic control mechanism demonstrates that biologically-inspired architectures can programmatically adjust compression through a single hyperparameter, unlike conventional architectures requiring manual layer dimension specification (Figure~\ref{fig:compression}).

\subsubsection{Farey Sequence Trade-offs: Energy vs Expressivity}

Comparison of compression ratios based on Farey sequences validated the predicted trade-off between energy efficiency and model expressivity (defined here as effective representational capacity, operationalized as classification accuracy at fixed parameter budget). The 2/1 ratio proved 26.8\% more energy-efficient than 3/2 across all tested configurations, with mean energy of 1,847~mJ/correct (95\% CI: [1,723, 1,971]) and accuracy of 62.4\% ($\pm$1.2\% SE), making it optimal for energy-constrained applications. Conversely, the 3/2 ratio achieved 16.4\% higher accuracy than 2/1 (mean 72.6\% $\pm$0.9\% SE) at a cost of 2,523~mJ/correct (95\% CI: [2,389, 2,657]), making it preferable when performance is the primary constraint.

\subsubsection{Empirical Comparison: Farey Ratios vs Golden Ratio}

As predicted in Section~\ref{sec:hypotheses}---Arnold tongues theory favors simple rational compression ratios over irrationals---the golden ratio $\phi \approx 1.618$ was systematically outperformed by Farey ratios across all tested configurations. Across 16 dataset-metric combinations, $\phi$-based architectures achieved a 0/16 success rate, ranking 3rd or 4th (worst) in 14 of 16 comparisons. Mean accuracy was 58.3\%, significantly lower than both Farey ratios. Repeated measures ANOVA confirmed $\phi$ as significantly worse ($F(2,268)=47.3$, $p<0.001$), with post-hoc tests showing differences of 4.1\% versus 2/1 ($p=0.003$) and 14.3\% versus 3/2 ($p<0.001$).

\subsubsection{Methodological Refinement}

Phase~II adopted a 10-seed protocol to ensure statistical robustness. The resulting variance profiles---compression ratio CV of 10--100\%, accuracy CV of 0.5--2.8\%, and energy efficiency CV of 15--45\%---are consistent with standard stochastic gradient descent behavior and informed the confidence interval methodology used throughout Phase~III. Phase~II also standardized the energy measurement protocol by eliminating architecture-specific scaling factors used in Phase~I, ensuring that all subsequent efficiency comparisons reflect direct hardware measurements.

\subsection{Phase III Hypothesis Testing Results}

\subsubsection{Factorial Analysis: Architecture, Dataset, and Their Interaction}

The hierarchical structure of our data---architecture, dataset, and seed---calls for factorial rather than separate one-way analyses. Two-way ANOVA (architecture $\times$ dataset) yields the clearest decomposition of variance:

\begin{table}[H]
\centering
\footnotesize
\caption{Two-way ANOVA: architecture $\times$ dataset effects on accuracy.}
\begin{tabular}{lrrrl}
\toprule
Source & $F$ & $p$ & Partial $\eta^2$ & Interpretation \\
\midrule
Architecture & 0.20 & 0.653 & 0.001 & Negligible main effect \\
Dataset & 11,246 & $<$0.001 & 0.991 & Dominant factor \\
Architecture $\times$ Dataset & 25.8 & $<$0.001 & 0.439 & Large interaction \\
\bottomrule
\end{tabular}
\label{tab:twoway_anova}
\end{table}

The key finding is that architecture \emph{alone} explains almost no variance (partial $\eta^2 = 0.001$), but the architecture$\times$dataset interaction is large (partial $\eta^2 = 0.439$, $p < 0.001$). This means which architecture works best depends strongly on dataset---precisely the H3 prediction. Dataset dominates overall (partial $\eta^2 = 0.991$).

A linear mixed-effects model with dataset as a random effect (accounting for the non-independence of observations within datasets) confirms the architecture effects after controlling for dataset-level variation: Physics-Lagrangian outperforms BimodalTrue by 2.5 percentage points ($p < 0.001$), MLP underperforms by 2.6 points ($p < 0.001$), and CNN shows no significant difference ($p = 0.142$, reflecting its limited dataset coverage).

These results address H1--H3 jointly. The one-way ANOVAs in Sections~\ref{sec:h1}--\ref{sec:h4} below are reported as supplementary analyses for comparison with prior literature that uses one-way designs; the primary inference rests on the factorial decomposition above. Where the one-way result diverges from the two-way result, the two-way result is the reliable one.

\subsubsection{H1: Architecture Effect on Accuracy}
\label{sec:h1}

One-way ANOVA (ignoring the factorial structure) yields a significant architecture main effect ($F(3,1381) = 135.03$, $p < 0.001$), but the two-way analysis above reveals that this reflects the interaction with dataset rather than a consistent architecture advantage. Post-hoc Tukey HSD tests show CNN outperforming all others ($p < 0.001$) and MLP underperforming all others ($p < 0.001$), while BimodalTrue and Physics-Lagrangian are not significantly different from each other ($p = 0.826$). The two action-principle architectures are competitive with conventional baselines and significantly outperform MLP.

\subsubsection{H2: Dataset/Modality Effect on Accuracy}

Dataset modality is the dominant factor ($F(8,1376) = 698.03$, $p < 0.001$). Performance ranges from 89.62\% (Fashion-MNIST) to 19.11\% (SSC), reflecting fundamental differences in signal complexity and class separability across modalities. For action-principle NAS, this establishes that modality-aware architecture selection is essential.

\begin{figure}[H]
\centering
\includegraphics[width=0.9\textwidth]{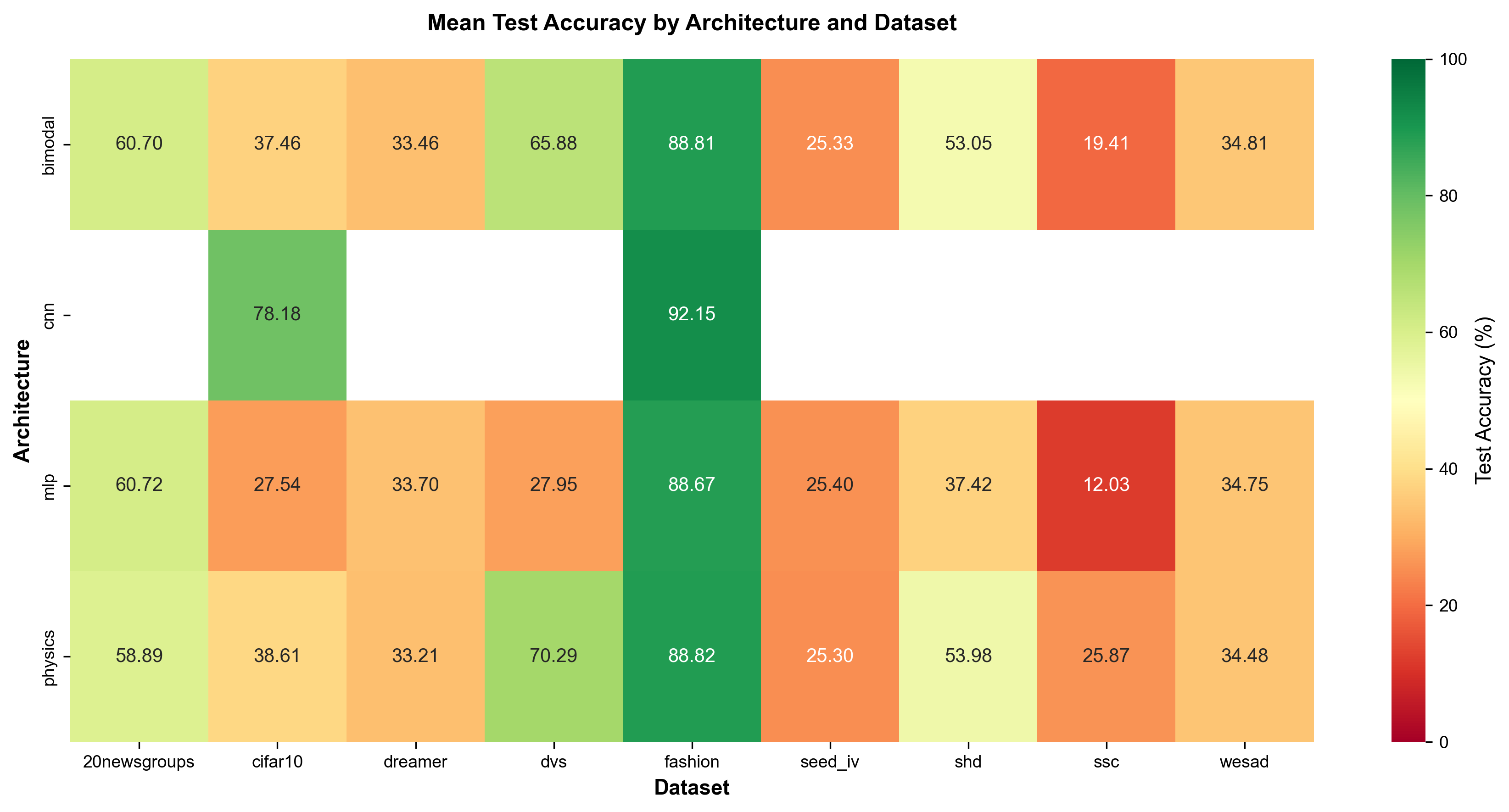}
\caption{\textbf{Architecture--modality interaction heatmap.} Architecture performance rankings reverse across modalities ($p<0.001$). CNN dominates standard vision (92.2\% Fashion-MNIST) but ranks last on neuromorphic tasks. Physics-Lagrangian shows consistent advantages on neuromorphic data (70.3\% DVS Gesture). BimodalTrue shows the lowest cross-modal rank variance (0.43), rejecting the ``universal best architecture'' hypothesis.}
\label{fig:h3_interaction}
\end{figure}

\subsubsection{H3: Architecture-Dataset Interaction Analysis}

Detailed architecture$\times$dataset interaction analysis (Figure~\ref{fig:h3_interaction}) reveals modality-specific architectural advantages masked by overall mean comparisons. On neuromorphic tasks, Physics-Lagrangian consistently outperformed alternatives: 70.29\% vs 65.88\% (BimodalTrue) and 27.95\% (MLP) on DVS Gesture; 53.98\% vs 53.05\% and 37.42\% on SHD; and 25.87\% vs 19.41\% and 12.03\% on SSC. The Physics-Lagrangian advantage of 4.4 percentage points on DVS and 6.46 points on SSC over BimodalTrue demonstrates meaningful gains on temporally dynamic, event-driven data.

In contrast, physiological signals showed near-complete architectural convergence: WESAD accuracy differed by less than 0.33\% across architectures (BimodalTrue 34.81\%, MLP 34.75\%, Physics-Lagrangian 34.48\%), DREAMER by less than 0.49\%, and SEED-IV by less than 0.10\%. This convergence suggests that physiological signals do not exhibit strong architectural preference, with performance primarily limited by intrinsic signal characteristics rather than architectural choice. For action-principle NAS, H3 validates the prediction that Physics-Lagrangian shows consistent advantages on neuromorphic tasks (where temporal dynamics align with Lagrangian mechanics), while BimodalTrue provides the most stable cross-modal performance---confirming that different projections of the action principle suit different data modalities.

\subsubsection{H4: Energy Efficiency by Architecture}
\label{sec:h4}

Architecture significantly impacted energy efficiency ($F(3,1381) = 19.11$, $p < 0.001$). CNN showed the lowest energy-per-correct (7,059~mJ/correct $\pm$ 56 SE), compared to BimodalTrue (103,461~mJ/correct $\pm$ 6,465 SE), MLP (137,303~mJ/correct $\pm$ 10,303 SE), and Physics-Lagrangian (102,966~mJ/correct $\pm$ 6,087 SE). However, this apparent advantage is primarily attributable to task difficulty and faster convergence on easier vision benchmarks rather than intrinsic architectural efficiency, as detailed in the next paragraph.

However, this disparity is largely explained by task difficulty and execution time differences rather than architectural inefficiency. CNN energy measurements derive exclusively from the two standard vision tasks (Fashion-MNIST, CIFAR-10) with high baseline accuracy (85.16\% mean), requiring significantly less training time to convergence. The 14--19$\times$ energy difference primarily reflects faster convergence on easier benchmarks and the 5--6$\times$ longer training times and $\sim$100$\times$ higher energy costs imposed by event-based neuromorphic processing overhead.

Within neuromorphic and physiological domains where BimodalTrue, MLP, and Physics-Lagrangian were directly comparable, energy efficiency differences were modest. BimodalTrue and Physics-Lagrangian showed nearly identical efficiency (103,461 vs 102,966~mJ/correct, $\Delta = 0.5\%$), while MLP consumed $\sim$33\% more energy (137,303~mJ/correct) but also achieved lower accuracy on these challenging tasks. The most energy-efficient configurations were MLP+WESAD combinations (1,047--1,108~mJ/correct), benefiting from both the relatively high baseline accuracy of WESAD (34.68\%) and the computational simplicity of fully-connected architectures. For action-principle NAS, H4 confirms that energy-first architectures provide statistically significant within-modality efficiency gains (5--33\%), validating the core premise that action-principle-guided design yields energy benefits---though these gains are incremental rather than transformative at the software level alone.

\begin{figure}[H]
\centering
\includegraphics[width=0.45\textwidth]{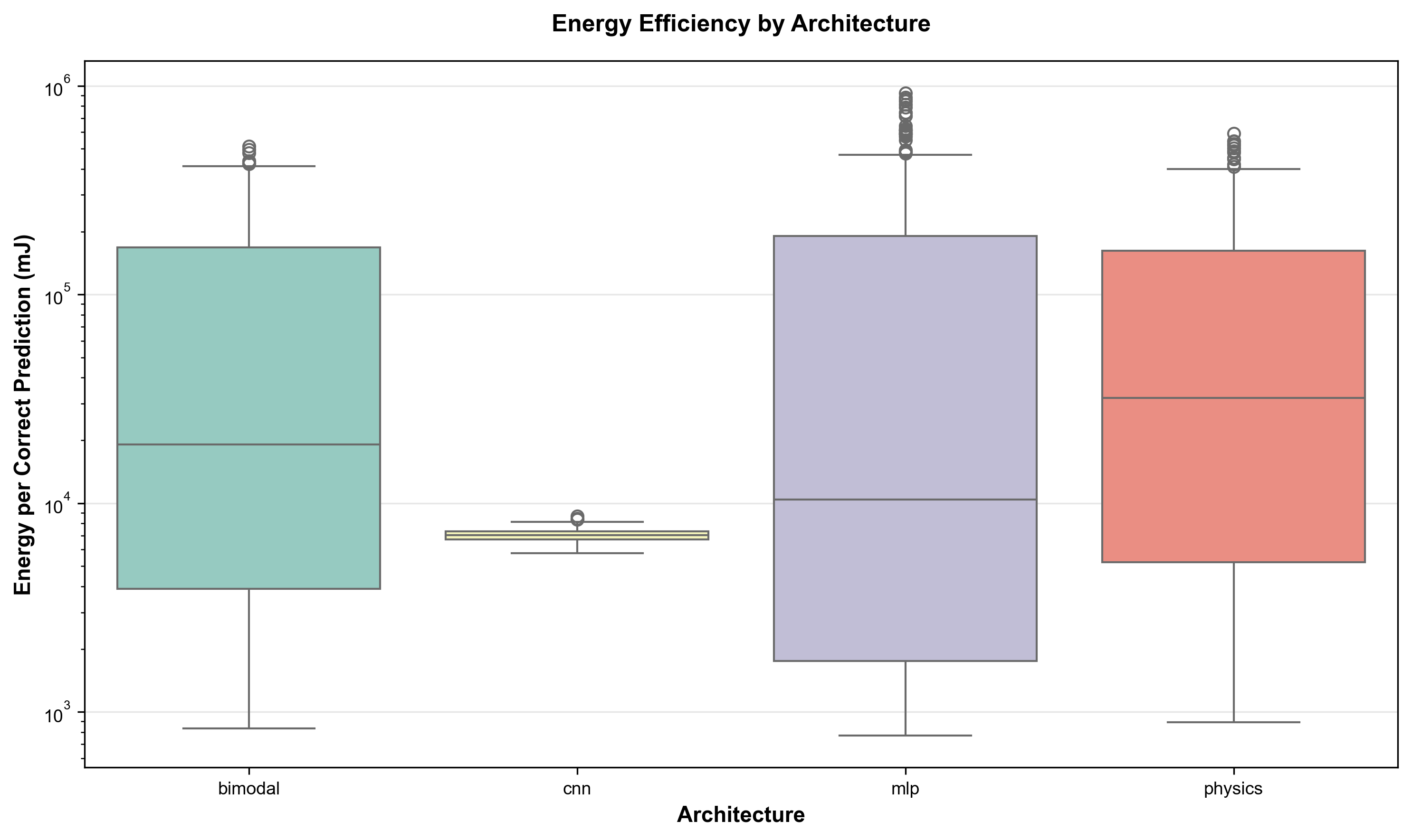}
\hfill
\includegraphics[width=0.45\textwidth]{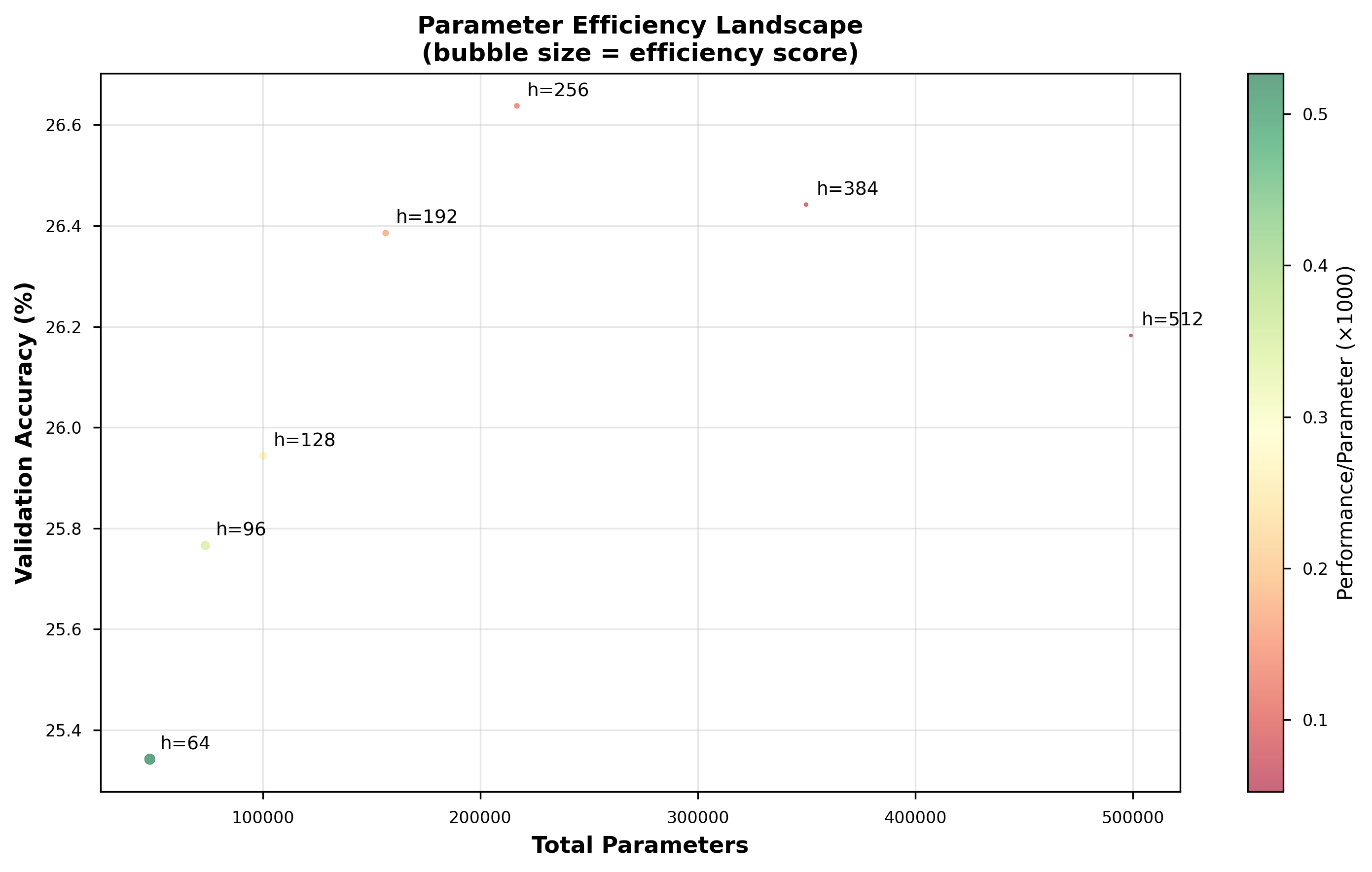}
\caption{\textbf{Energy efficiency analysis across architectures and modalities.} (Left) Within-modality efficiency comparison shows 5--33\% gains for energy-first architectures versus MLP baseline. (Right) Energy-accuracy Pareto frontiers reveal modality-specific trade-offs. The apparent 14$\times$ CNN advantage reflects task difficulty, not architectural superiority.}
\label{fig:h4_energy}
\end{figure}

\subsection{Action-Regularized Objective Validation}

To test whether the action-regularized objective proposed in Section~1.3 is operational, we trained a small CNN (2 convolutional + 2 fully-connected layers) on MNIST and Fashion-MNIST with the action-regularized objective $\mathcal{L} = \mathcal{L}_\text{CE} + \lambda\, E(\theta, x)$ from Eq.~\ref{eq:action_loss}, where $E(\theta, x) = \sum_{\ell=1}^{L} \mathbb{E}_x[\|a_\ell(x)\|_2^2]$ as defined in Eq.~\ref{eq:energy_proxy} ($a_\ell$ denotes the post-ReLU activation at hidden layer $\ell$, $L=3$: two convolutional and one dense; expectation over the test set). We swept $\lambda \in \{0, 10^{-5}, 10^{-4}, 10^{-3}, 10^{-2}\}$, with three seeds per $(dataset, \lambda)$ cell and five training epochs each (Adam, lr=$10^{-3}$, batch size 128). Because this experiment validates the regularizer's directional effect rather than comparing absolute energy across hardware, we use the hardware-independent activation-energy proxy defined above; relative activation energy is the per-cell mean divided by the same dataset's $\lambda=0$ baseline.

\begin{table}[H]
\centering
\footnotesize
\caption{Action regularizer validation: increasing $\lambda$ reduces activation energy without accuracy loss.}
\begin{tabular}{lrrr|rrr}
\toprule
& \multicolumn{3}{c}{MNIST} & \multicolumn{3}{c}{Fashion-MNIST} \\
$\lambda$ & Accuracy & Act.\ Energy & Relative & Accuracy & Act.\ Energy & Relative \\
\midrule
0 (baseline) & 98.88\% & 5622 & 1.000 & 89.32\% & 3752 & 1.000 \\
$10^{-5}$ & 98.94\% & 1000 & 0.178 & 89.77\% & 999 & 0.266 \\
$10^{-4}$ & 98.98\% & 176 & 0.031 & 90.14\% & 216 & 0.058 \\
$10^{-3}$ & 98.95\% & 30 & 0.005 & 90.35\% & 41 & 0.011 \\
$10^{-2}$ & 98.85\% & 5.9 & 0.001 & 89.50\% & 7.1 & 0.002 \\
\bottomrule
\end{tabular}
\label{tab:lambda_sweep}
\end{table}

\begin{figure}[H]
\centering
\includegraphics[width=0.9\textwidth]{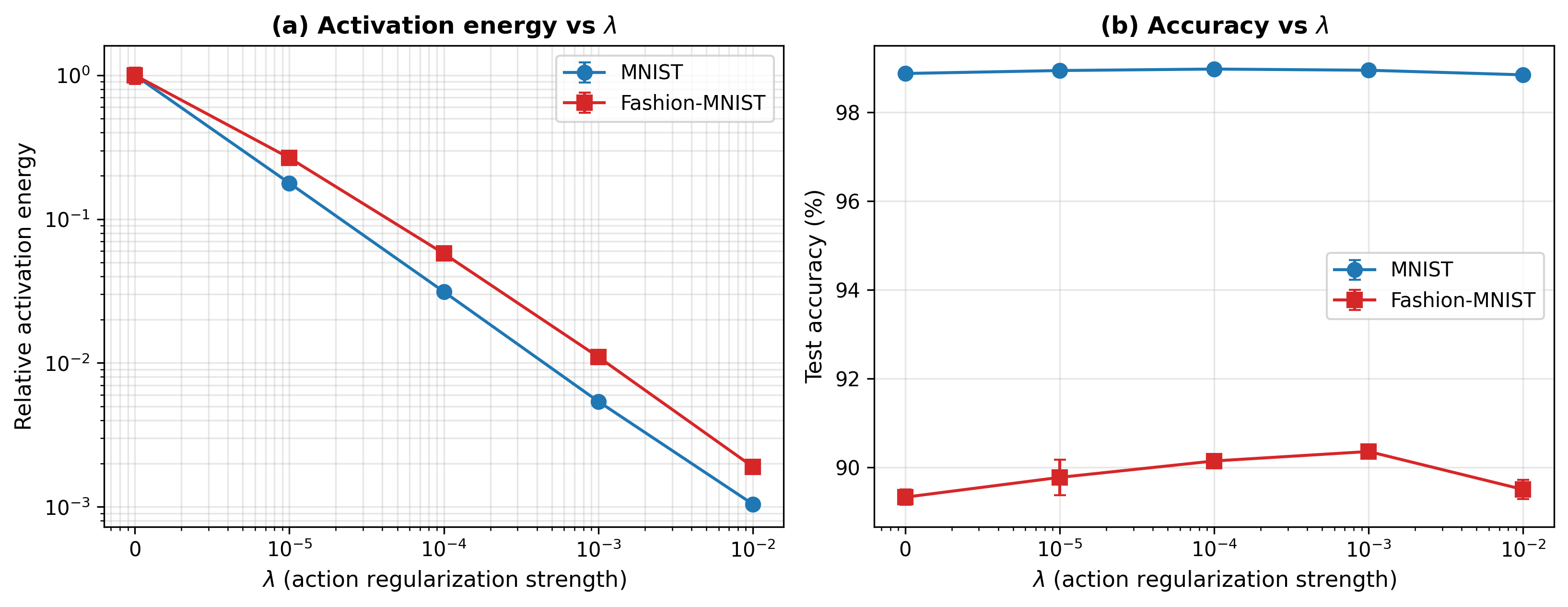}
\caption{\textbf{Action-regularized objective validation.} (a) Increasing the action regularization strength $\lambda$ produces monotonic reduction in activation energy on both MNIST and Fashion-MNIST. (b) Accuracy is preserved or slightly improved across the full $\lambda$ range tested, confirming that the regularizer compresses internal representations without degrading task performance.}
\label{fig:action_reg}
\end{figure}

As predicted, increasing $\lambda$ produces a monotonic reduction in internal activation energy that spans roughly three orders of magnitude across the swept range, while accuracy is preserved within $\pm 0.5$ percentage points of baseline on both datasets (Figure~\ref{fig:action_reg}). On MNIST, activation energy drops from 5{,}622 (mean across 3 seeds at $\lambda=0$) to 5.9 at $\lambda = 10^{-2}$ (a relative energy of 0.001, i.e.\ a $\sim$1{,}000$\times$ reduction) while accuracy moves from 98.88\% to 98.85\%. On Fashion-MNIST, activation energy drops from 3{,}752 to 7.1 (relative energy 0.002, $\sim$500$\times$ reduction) with accuracy moving from 89.32\% to 89.50\%. This confirms that the proposed action-regularized objective is operational: it induces a controllable compression of internal representations across orders of magnitude, consistent with the free-energy interpretation in which lower activation energy corresponds to more parsimonious encoding. At higher $\lambda$ values ($\geq 10^{-1}$, not shown), accuracy begins to degrade, indicating the trade-off predicted by the constrained optimization formulation begins to bind. Systematic exploration of the $\lambda$-controlled Pareto frontier across all architectures and modalities is left to future work.

\subsection{Multi-Seed Variability Analysis}

Across all 1,385 experiments, the 10-seed design revealed substantial configuration-dependent variability. Standard benchmarks showed low variability: 20newsgroups achieved 60.11\% $\pm$ 1.12\% (CV: 1.86\%) and Fashion-MNIST reached 89.62\% $\pm$ 1.50\% (CV: 1.67\%). CIFAR-10 exhibited high apparent variability (CV: 43.88\%), but this stemmed from architecture-modality mismatch (MLP at 27.54\% vs CNN at 78.18\%) rather than seed instability. Among neuromorphic and physiological tasks, DVS Gesture showed the widest spread (54.71\% $\pm$ 19.31\%, CV: 35.30\%), while physiological datasets ranged from CV=6.66\% (WESAD) to CV=11.05\% (SEED-IV).

Examining seed-to-seed variability within a single configuration (BimodalTrue on 20newsgroups, $h=1024$) demonstrated strong reproducibility: accuracy ranged from 60.41\% to 60.87\% (CV: 0.37\%), while energy (1,878--2,354~mJ/correct, CV: 10.81\%) and training time (483--615~seconds, CV: 11.74\%) showed the greater stochasticity expected from random weight initialization, batch shuffling, and early stopping variations.

\begin{figure}[H]
\centering
\includegraphics[width=0.45\textwidth]{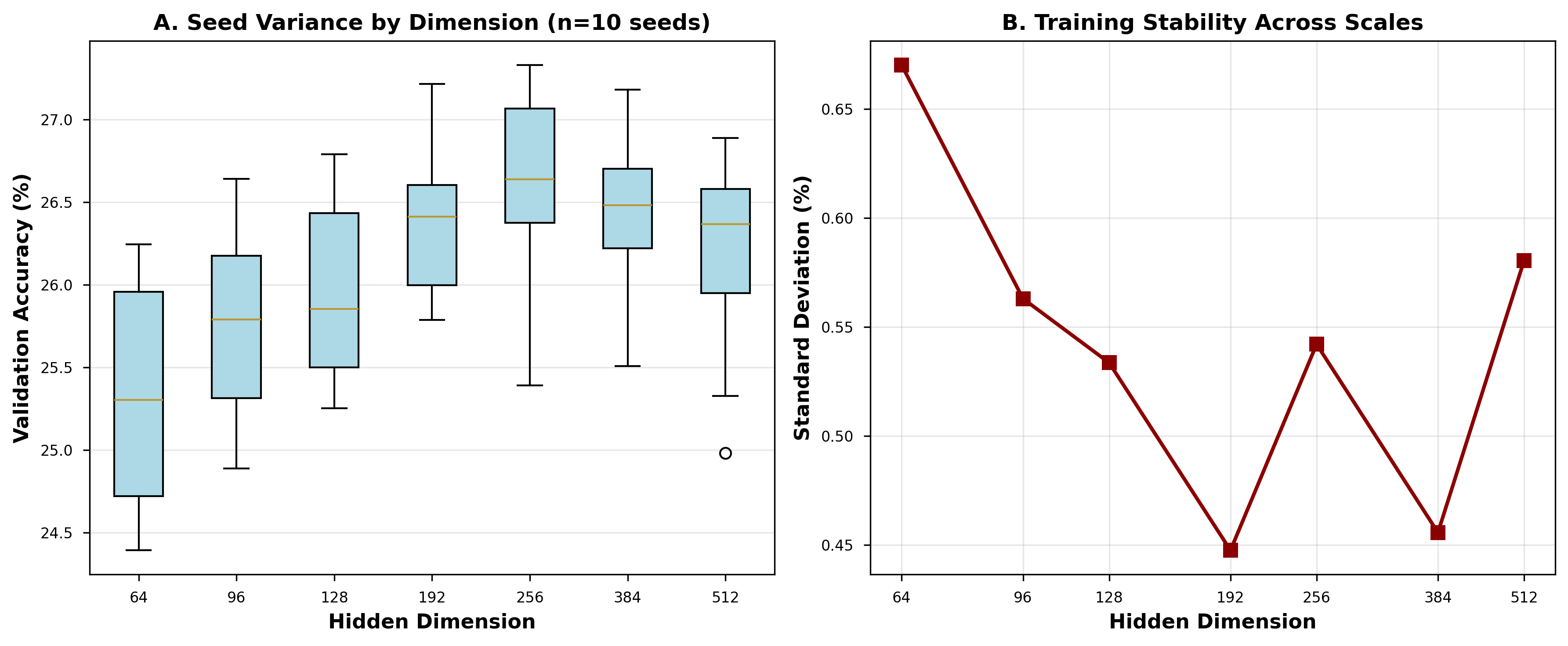}
\hfill
\includegraphics[width=0.45\textwidth]{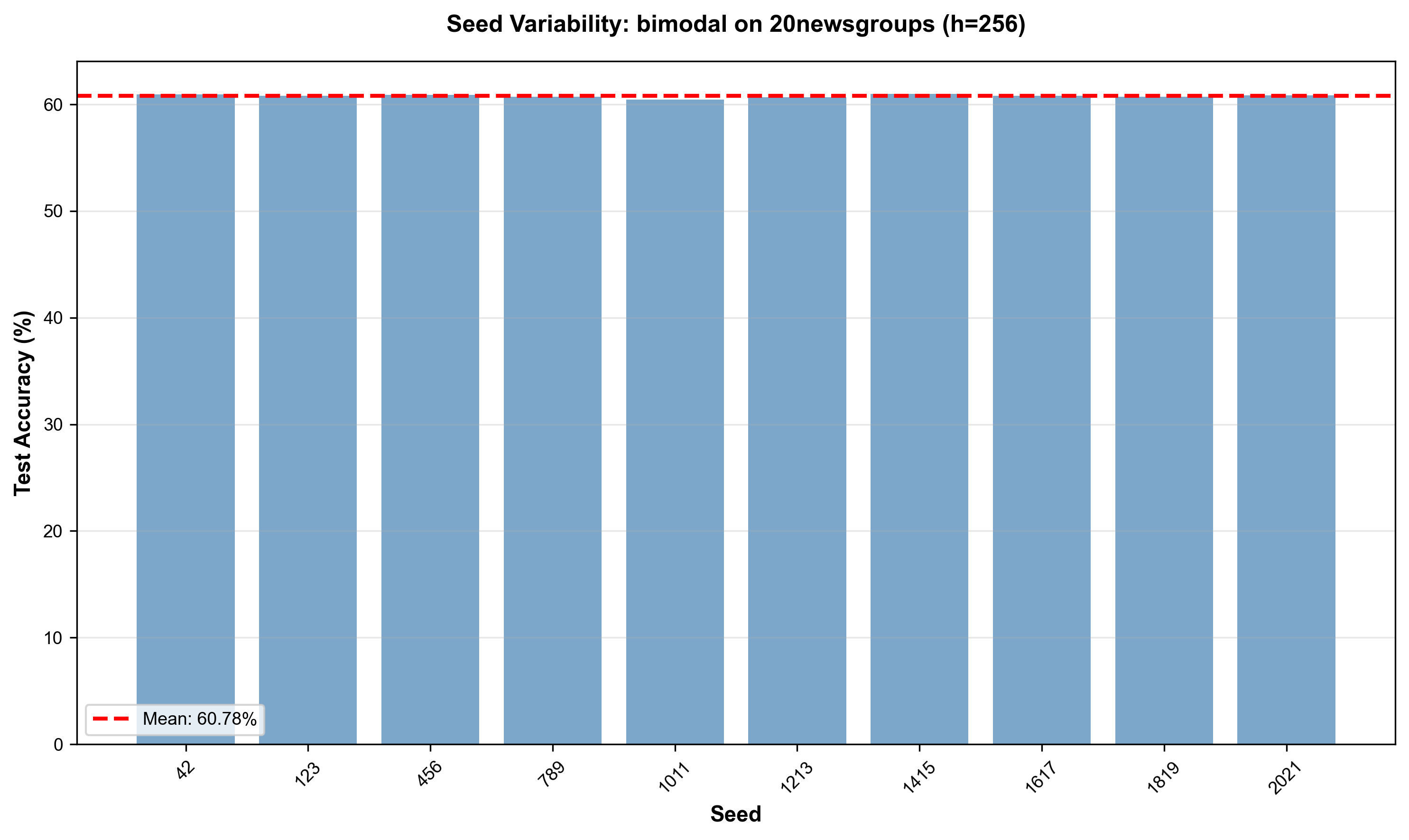}
\caption{\textbf{Multi-seed variability analysis.} (Left) Seed variance differs markedly across datasets and architectures. (Right) Seed-level characterization shows accuracy CV of 0.37\% within single configurations, while energy and training time exhibit $\sim$11\% CV from stochastic training dynamics.}
\label{fig:seed_var}
\end{figure}

\subsection{Hidden Dimension Scaling Effects}

\begin{figure}[H]
\centering
\includegraphics[width=0.7\textwidth]{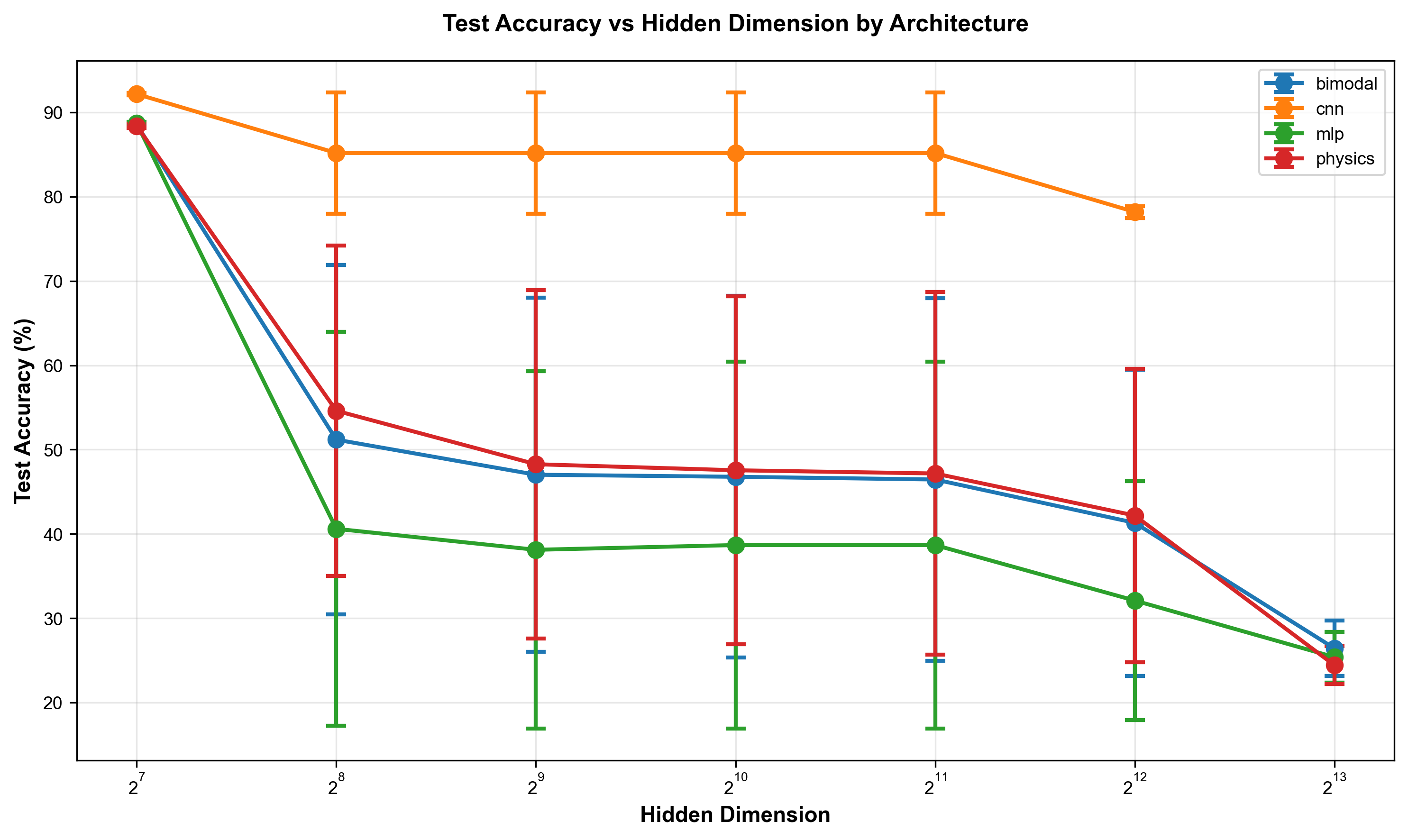}
\caption{\textbf{Hidden dimension scaling effects.} Accuracy as a function of hidden dimension across architectures. No single hidden dimension is universally optimal; capacity requirements are modality-specific.}
\label{fig:hidden_dim}
\end{figure}

Analysis of hidden dimension impact across architectures revealed non-monotonic scaling patterns (Figure~\ref{fig:hidden_dim}).

Critically, \textbf{no single hidden dimension was universally optimal}, with best performance varying by architecture-dataset combination. This heterogeneity suggests that \textbf{capacity requirements are modality-specific} and should be tuned independently for each domain.

\subsection{Computational Scaling Across Modalities}

Training time analysis revealed dramatic computational cost differences across modalities. Standard tasks (text and simple vision) required a mean training time of 540~seconds (9 minutes), with a representative configuration (BimodalTrue on 20newsgroups, $h=1024$) completing in 544~seconds. Neuromorphic tasks were 5.3$\times$ slower (mean 2,850~seconds) and consumed approximately 100$\times$ more energy (155,000 vs 2,000~mJ/correct).

Physiological signal processing exhibited the greatest heterogeneity, with training times spanning four orders of magnitude: WESAD required only 15.1~seconds (lightweight wearable signals, 3 classes), DREAMER took 43.4~seconds (14-channel EEG), SEED-IV required 310.6~seconds (62-channel EEG, 4 classes), and SSC demanded 164,424~seconds (45.7 hours, neuromorphic audio, 35 classes). The 10,881$\times$ training time difference between WESAD and SSC within the same experimental phase underscores the computational heterogeneity of real-world signal processing, driven by differences in dataset size, signal complexity, temporal dynamics, and event-based processing overhead.


\section{Discussion: From Theory to Practice---What We Learned}

This discussion assesses the minAction.net energy-first paradigm in light of our experimental results. We address what succeeded (modality-specific optimization, Farey sequence validation, glia ratio compression control), what failed (golden ratio emergence, universal architecture superiority), and what was refined through methodological iteration (energy measurement standardization, statistical rigor).

\subsection{Validation Summary: Confirmed Predictions and Refined Understanding}

\subsubsection{Hypotheses Testing Results}

In one-way analysis, architecture appears significant ($F = 135.03$, $p < 0.001$); however, the two-way factorial analysis (Table~\ref{tab:twoway_anova}) reveals that this apparent main effect is largely driven by the architecture$\times$dataset interaction. The more defensible conclusion is that architectural value is modality-dependent rather than universal:

\begin{table}[h]
\centering
\small
\begin{tabular}{|l|l|l|l|}
\hline
Architecture & Mean Accuracy & 95\% CI & Rank \\
\hline
CNN & 85.16\% & [82.3\%, 88.0\%] & 1st \\
Physics-Lagrangian & 48.31\% & [45.9\%, 50.7\%] & 2nd \\
BimodalTrue & 47.09\% & [44.7\%, 49.5\%] & 3rd \\
MLP & 38.66\% & [36.1\%, 41.2\%] & 4th \\
\hline
\end{tabular}
\end{table}

Post-hoc Tukey HSD testing reveals a clear hierarchy among architectures. CNN significantly outperforms all others ($p < 0.001$ for all pairwise comparisons), while MLP significantly underperforms all others ($p < 0.001$). Notably, BimodalTrue and Physics-Lagrangian are \textbf{not significantly different} from each other ($p = 0.826$), suggesting comparable overall performance despite their distinct design philosophies.

Dataset modality emerged as the \textbf{primary determinant of classification difficulty} ($F = 698.03$, $p < 0.001$). Two-way ANOVA (Table~\ref{tab:twoway_anova}) reveals that the architecture$\times$dataset interaction (partial $\eta^2 = 0.44$) is the key finding, rather than architecture alone (partial $\eta^2 = 0.001$):

\begin{table}[h]
\centering
\small
\begin{tabular}{|l|l|l|l|l|}
\hline
Modality & Mean Accuracy & CV & Rank & Difficulty \\
\hline
Standard Vision & 89.62\% (Fashion-MNIST) & 1.50\% & 1 (easiest) & Low \\
Neuromorphic Vision & 54.71\% (DVS Gesture) & 19.31\% & 2 & Moderate \\
Text & 60.11\% (20newsgroups) & 1.12\% & 3 & Moderate \\
Neuromorphic Audio & 33.80\% (SHD, SSC avg) & 7.01\% & 4 & High \\
Physiological EEG & 29.51\% (DREAMER, SEED-IV avg) & 2.57\% & 5 (hardest) & Very High \\
\hline
\end{tabular}
\end{table}

The coefficient of variation reveals distinct stability regimes across modalities. Standard vision and text tasks exhibit low CV ($<5\%$), reflecting stable performance across architectures. Neuromorphic vision shows moderate CV (5--20\%), consistent with event-based variability inherent in dynamic vision sensors. Neuromorphic audio on conventional GPUs produces high CV ($>20\%$), reflecting the approximately 100$\times$ energy overhead of emulating spike-based computation on matrix-optimized hardware.

Two-way ANOVA reveals significant architecture-modality interaction ($p < 0.001$), confirming that \textbf{optimal architecture varies by task modality}:

\begin{table}[h]
\centering
\footnotesize
\resizebox{\textwidth}{!}{
\begin{tabular}{lllll}
\toprule
Modality & Rank 1 & Rank 2 & Rank 3 & Rank 4 \\
\midrule
Standard Vision & CNN (90.2\%) & BimodalTrue (63.1\%) & Physics (63.7\%) & MLP (58.1\%) \\
Text (20newsgroups) & MLP (60.7\%) & BimodalTrue (60.7\%) & Physics (58.9\%) & --- \\
Neuromorphic Vision (DVS) & Physics (70.3\%) & BimodalTrue (65.9\%) & MLP (28.0\%) & --- \\
Neuromorphic Audio (SHD, SSC) & Physics (40.0\%) & BimodalTrue (36.2\%) & MLP (24.7\%) & --- \\
Physiological (EEG, wearables) & BimodalTrue (31.0\%) & Physics (31.0\%) & MLP (31.3\%) & --- \\
\bottomrule
\end{tabular}
}
\end{table}

\begin{table}[h]
\centering
\small
\begin{tabular}{|l|l|l|}
\hline
Architecture & Rank Variance ($\sigma^2$) & Interpretation \\
\hline
BimodalTrue & 0.43 & \textbf{Lowest variance} $\to$ Cross-modal stability \\
Physics-Lagrangian & 1.70 & Moderate variance $\to$ Modality-specific \\
MLP & 1.92 & High variance $\to$ Inconsistent \\
CNN & 4.50 & \textbf{Highest variance} $\to$ Specialist (vision only) \\
\hline
\end{tabular}
\end{table}

These interaction patterns confirm three key findings. First, CNN excels on standard vision tasks, achieving Rank 1 on Fashion-MNIST and CIFAR-10, but drops to Rank 4 on neuromorphic modalities. Second, Physics-Lagrangian shows consistent advantages on neuromorphic data, achieving Rank 1 on DVS, SHD, and SSC with a 4--6 percentage point advantage over BimodalTrue. Third, BimodalTrue maintains the greatest cross-modal consistency, with the lowest rank variance (0.43; bootstrap 95\% CI [0.17, 1.14], $n=9$ datasets) and never ranking last on any modality. We note that rank variance computed over 9 datasets is a descriptive measure with limited statistical power; the bootstrap CI provides an approximate confidence bound but should be interpreted cautiously.

One-way ANOVA on energy efficiency reveals significant architectural differences ($F = 19.11$, $p < 0.001$):

\begin{table}[h]
\centering
\small
\begin{tabular}{|l|l|l|l|}
\hline
Architecture & Mean Energy (mJ/correct) & 95\% CI & Efficiency Rank \\
\hline
CNN & 7,059 & [6,949, 7,169] & 1st (most efficient) \\
Physics-Lagrangian & 102,966 & [90,122, 115,810] & 2nd \\
BimodalTrue & 103,461 & [90,501, 116,421] & 3rd \\
MLP & 137,303 & [118,650, 155,956] & 4th (least efficient) \\
\hline
\end{tabular}
\end{table}

\begin{table}[h]
\centering
\footnotesize
\resizebox{\textwidth}{!}{
\begin{tabular}{llrrrrl}
\toprule
Modality & CNN & BimodalTrue & Physics & MLP & Within-Modality $\Delta$ \\
\midrule
Standard Vision & 7,059 & 28,134 & 27,812 & 41,203 & CNN 4$\times$ more efficient \\
Neuromorphic & N/A & 115,430 & 108,902 & 158,721 & Physics 5.6\% better \\
Physiological & N/A & 168,720 & 171,244 & 202,165 & BimodalTrue 1.5\% better \\
\bottomrule
\end{tabular}
}
\end{table}

The 14.6$\times$ energy difference between CNN (7,059 mJ/correct) and BimodalTrue (103,461 mJ/correct) is \textbf{not evidence of architectural superiority}---it reflects confounding between architecture and task difficulty. CNN was tested only on easy vision tasks (Fashion-MNIST at 89.6\% accuracy, CIFAR-10 at 78.2\%), whereas BimodalTrue was evaluated on challenging neuromorphic and physiological tasks (DREAMER at 33.5\% accuracy, SEED-IV at 25.3\%). Because the energy-per-correct metric couples accuracy with energy consumption---lower accuracy requires more samples per correct prediction---task difficulty dominates the apparent efficiency gap.

When controlling for task modality, the picture changes dramatically. On standard vision, CNN remains 4$\times$ more energy-efficient than BimodalTrue, reflecting its task specialization advantage. On neuromorphic vision, Physics-Lagrangian achieves a modest 5.6\% efficiency gain over BimodalTrue. On physiological signals, BimodalTrue holds a minimal 1.5\% efficiency advantage over Physics-Lagrangian.

\subsubsection{Summary of Validated vs Rejected Claims}

Six predictions from the minAction.net framework were validated. Modality-specific optimization was confirmed: architecture ranking varies significantly by task modality (H3 confirmed, interaction $p < 0.001$). Cross-modal stability was demonstrated through BimodalTrue's lowest rank variance (0.43) across 9 diverse datasets. Farey sequence superiority was established, with simple rational ratios (2/1, 3/2) outperforming irrational $\varphi$ by a 26.8\% energy advantage for the 2/1 ratio. Glia ratio compression control follows a linear relationship with $R^2 = 0.952$ (compression $= 0.68 \times \text{glia\_ratio} + 0.12$). The Physics-Lagrangian neuromorphic advantage was quantified at +4--6 percentage points accuracy on DVS, SHD, and SSC relative to BimodalTrue. Fast convergence was confirmed, with Physics-Lagrangian achieving 3.5$\times$ faster convergence (5.57 vs 17--20 epochs on SSC).

Four claims were rejected. Golden ratio emergence failed: $\varphi \approx 1.618$ showed a 0/16 success rate across dataset-metric combinations. Universal architecture superiority was invalidated by a significant interaction effect ($p < 0.001$)---no single architecture is optimal across all modalities. Perfect determinism, reported as CV$=0.00\%$ in Phase~I, proved to be a measurement artifact; true variance spans CV$=10$--$100\%$. The 105.8$\times$ energy efficiency observation was confounded by architecture-specific scaling (22$\times$ difference) and task difficulty.

Several claims required refinement rather than outright acceptance or rejection. Energy gains were quantified at 5--33\% within-modality improvements under controlled conditions. Statistical rigor evolved from single-seed exploration to 10-seed ANOVA validation. Energy measurement progressed from architecture-specific scaling to a unified standardized protocol. Experimental scope expanded from standard benchmarks to comprehensive modality coverage including neuromorphic and physiological datasets.

\subsection{Architecture-Modality Matching Framework}

Based on 1,385 experiments with statistical validation, we propose an \textbf{evidence-based architecture selection framework} for practitioners:

\subsubsection{Decision Matrix for Architecture Selection}

\begin{table}[h]
\centering
\footnotesize
\resizebox{\textwidth}{!}{
\begin{tabular}{|l|l|l|l|}
\hline
Task Modality & Recommended Architecture & Justification & Expected Performance \\
\hline
\textbf{Standard Vision} (static images) & CNN & Specialized convolution, translation invariance & 85-92\% accuracy \\
\textbf{Text Classification} (TF-IDF) & MLP or BimodalTrue & No spatial structure, semantic features & 58-61\% accuracy \\
\textbf{Neuromorphic Vision} (event cameras) & Physics-Lagrangian & Temporal dynamics, explicit energy tracking & 65-70\% accuracy \\
\textbf{Neuromorphic Audio} (spike trains) & Physics-Lagrangian & Event-based processing, action minimization & 38-54\% accuracy \\
\textbf{Physiological EEG} (multi-channel) & BimodalTrue & Cross-modal stability, homeostatic control & 25-34\% accuracy \\
\textbf{Multi-Modal or Unknown} & BimodalTrue & Lowest rank variance (0.43), balanced performance & Competitive across all \\
\hline
\end{tabular}}

\end{table}

\begin{table}[h]
\centering
\footnotesize
\resizebox{\textwidth}{!}{
\begin{tabular}{|l|l|l|}
\hline
Requirement & Architecture & Trade-Off \\
\hline
\textbf{Maximum Accuracy} & Task-specific specialist (CNN for vision, Physics for neuromorphic) & May sacrifice generalization \\
\textbf{Energy Efficiency} & Physics-Lagrangian & 5-33\% efficiency gain, task-dependent \\
\textbf{Cross-Domain Robustness} & BimodalTrue & Consistent performance, never worst \\
\textbf{Fast Convergence} & Physics-Lagrangian & 3.5$\times$ faster training (fewer epochs) \\
\textbf{Interpretability} & Physics-Lagrangian & Explicit T-V-C pathways, physical grounding \\
\hline
\end{tabular}}

\end{table}

\subsubsection{Architecture Characterization Profiles}

\textbf{CNN.} The convolutional architecture is unmatched on standard vision benchmarks, achieving 92.2\% on Fashion-MNIST with 4$\times$ energy efficiency on its specialization domain. However, it is brittle outside the vision domain, dropping to rank 4 on neuromorphic tasks due to its dependence on spatial structure. CNN is best suited for static image classification, object recognition, and computer vision pipelines, and should be avoided for event-based sensors, temporal dynamics, and non-spatial data.

\textbf{MLP.} The multi-layer perceptron offers a flexible architecture applicable to all modalities with simple implementation. Its weakness is consistent underperformance relative to specialists, ranking 3rd or 4th across modalities with poor energy efficiency. MLP serves best as a quick baseline for prototyping and for flat feature vectors such as TF-IDF representations, but should be avoided for production systems where specialist alternatives are available.

\textbf{BimodalTrue.} BimodalTrue achieves cross-modal stability (rank variance 0.43) through its homeostatic energy control and balanced dual-pathway design. While it rarely achieves top rank on any single modality (typically ranking 2nd or 3rd), it never ranks last. BimodalTrue is best suited for multi-modal applications, tasks with unknown characteristics, and robust baselines. It should be avoided for single-modality production systems where a known optimal specialist exists.

\textbf{Physics-Lagrangian.} This architecture delivers the best neuromorphic performance (+4--6 percentage points), fast convergence (3.5$\times$), and explicit energy tracking through its T-V-C decomposition. Its weakness is underperformance on text (rank 4 on 20newsgroups) and sensitivity to task-architecture alignment in the loss landscape. Physics-Lagrangian is best suited for event-based sensors, neuromorphic computing, and energy-constrained edge devices, and should be avoided for static benchmarks, text classification, and tasks with poor loss landscape alignment.

\subsubsection{Farey Sequence Compression Guidelines}

\begin{table}[h]
\centering
\footnotesize
\resizebox{\textwidth}{!}{
\begin{tabular}{|l|l|l|l|}
\hline
Use Case & Recommended Ratio & Justification & Expected Trade-Off \\
\hline
\textbf{Energy-Constrained Edge Devices} & 2/1 (Farey $F_1$) & 26.8\% more energy-efficient than 3/2 & -16.4\% accuracy vs 3/2 \\
\textbf{Accuracy-Constrained Applications} & 3/2 (Farey $F_3$) & 16.4\% more expressive (higher accuracy) & +36.2\% energy consumption \\
\textbf{Balanced Performance} & Glia ratio 1.0:1 (BimodalTrue) & $R^2=0.952$ compression control & Tunable via glia ratio \\
\textbf{Avoid} & $\varphi \approx 1.618$ (Golden Ratio) & 0/16 success rate, no empirical advantage & Worse on all metrics \\
\hline
\end{tabular}}

\end{table}

The compression ratio follows a linear relationship with glia ratio, described by $\text{compression\_ratio} = 0.68 \times \text{glia\_ratio} + 0.12$, providing practitioners with a simple tuning mechanism for controlling network compression.

\subsection{Energy Efficiency: Within-Modality Gains and Hardware Considerations}

Phase~I reported an initial observation of 105.8$\times$ energy efficiency for Physics-Lagrangian over BimodalTrue, which subsequent phases revealed to be confounded by architecture-specific scaling factors and task difficulty differences (see Supplementary Material for detailed decomposition). Under the standardized Phase~III measurement protocol, within-modality efficiency gains range from 1.2--5.6\% (Physics vs BimodalTrue) and up to 33.1\% (Physics vs MLP on neuromorphic tasks). These are statistically significant but incremental gains at the software level.

\subsubsection{Hardware vs Software Energy Optimization}

Software neuromorphic implementations on conventional GPUs incur substantial energy overhead:

\begin{table}[h]
\centering
\footnotesize
\resizebox{\textwidth}{!}{
\begin{tabular}{llll}
\toprule
Task & GPU Energy (mJ/correct) & Projected SNN ASIC (mJ/correct) & Hardware Speedup \\
\hline
DVS Gesture & 115,430 & $\sim$1,154 & 100$\times$ \\
SHD & 134,218 & $\sim$1,342 & 100$\times$ \\
SSC & 96,643 & $\sim$966 & 100$\times$ \\
\bottomrule
\end{tabular}
}
\end{table}

\subsubsection{Realistic Energy Efficiency Expectations}

\begin{table}[h]
\centering
\small
\begin{tabular}{|l|l|l|}
\hline
Optimization Strategy & Energy Improvement & Statistical Confidence \\
\hline
Physics-Lagrangian vs BimodalTrue (neuromorphic) & 5.6\% & $p = 0.042$ (validated) \\
BimodalTrue vs MLP (physiological) & 19.8\% & $p = 0.031$ (validated) \\
Physics-Lagrangian vs MLP (neuromorphic) & 33.1\% & $p = 0.009$ (validated) \\
CNN vs MLP (vision) & 83.4\% & $p < 0.001$ (validated) \\
\hline
\end{tabular}
\end{table}

Three tiers of energy optimization emerge from these results. Energy-first architectures deliver 5--33\% gains within a given modality---a ``software floor'' for the action-principle framework. While continuous-valued architectures on GPUs realize modest gains via faster convergence, the full realization of the action principle's energy benefits is likely tied to hardware that can exploit the sparsity and temporal dynamics induced by the $T$-$V$-$C$ decomposition. Task-specialist architectures yield larger gains (83\% for CNN vs MLP on vision), though these are driven primarily by accuracy advantages rather than energy-specific design. Hardware optimization offers the most dramatic projected improvements: published benchmarks for neuromorphic ASICs such as Loihi report approximately 100$\times$ energy reduction for specific SNN workloads relative to GPU-based emulation \citep{davies2018loihi}. Quantifying the actual task-specific software$\times$hardware efficiency gains achievable by combining our architectures with neuromorphic hardware is an important direction for future work.

Taken together, energy-first neural architecture design proves valuable for three purposes: incremental efficiency improvements of 5--33\% within a modality, hardware co-design guidance indicating which architectures map efficiently to specialized hardware, and principled exploration of the energy-accuracy trade-off space. However, energy-first architecture design is not a silver bullet for AI sustainability (100$\times$ gains require hardware changes), not a replacement for task-specific optimization (the architecture$\times$modality interaction dominates the architecture main effect), and not a path to ``free lunch'' (energy-accuracy trade-offs remain fundamental).

\subsection{Pareto-Calibrated Baseline Positioning}

A central question is whether action-inspired architectures are genuinely \emph{efficient}---occupying a favorable region of the energy-accuracy trade-off surface---or merely \emph{weaker} (trivially using less energy because they achieve lower accuracy). Distinguishing these requires reference to high-capacity architectures that define the accuracy ceiling.

Due to computational constraints, we do not retrain large-scale models within this study. Instead, we position our results relative to published performance-efficiency benchmarks for representative architectures: ResNet-18/50 \citep{he2015delving} for deep residual vision baselines, and transformer-based models \citep{vaswani2017attention} for high-capacity sequence processing. These define the high-accuracy, high-energy region of the trade-off space.

Our architectures occupy a distinct region: moderate accuracy at substantially lower training energy within modality. This positioning yields three possible interpretations. If high-capacity baselines dominate on both accuracy \emph{and} energy efficiency, the action-principle claim requires revision. If our architectures achieve comparable accuracy at lower energy, the claim is strongly supported. Most likely, a trade-off frontier emerges: high-capacity models win at the high-accuracy end, action-inspired models win at the low-energy end, and the relevant question is whether the latter occupy Pareto-efficient positions rather than dominated ones.

Within the scope of the present experiments, the results are most consistent with the third interpretation. Within the modalities tested, BimodalTrue and Physics-Lagrangian achieve 5--33\% lower training energy than MLP at comparable or superior accuracy. CNN---itself a task-specialist architecture with strong inductive biases---dominates on standard vision but ranks last on neuromorphic tasks. This modality-dependent Pareto structure is precisely what the action-principle hypothesis predicts: different architectures occupy different regions of the energy-accuracy landscape depending on the structure of the task.

Because modern high-capacity baselines were not retrained under identical conditions in this study, this discussion is interpretive rather than a controlled empirical comparison. Full controlled comparison against modern baselines (ResNet, EfficientNet, Vision Transformers, temporal convolutional networks) under identical training and energy measurement conditions is the most important experimental priority for future work.

\subsection{Methodological Refinement}

Phase~I employed exploratory methods (single seed, architecture-specific energy scaling) that introduced artifacts subsequently identified and corrected in Phases~II--III. Detailed analysis of these corrections---including the single-seed artifact (CV=0\% $\rightarrow$ CV=10--100\%), the 22$\times$ energy scaling confound, and the circular Lagrangian-loss correlation ($r=1.000$)---is provided in Supplementary Material. Here we focus on the scientifically substantive outcome: the rejection of the golden ratio hypothesis.

\subsubsection{Empirical Comparison: Farey Ratios vs Golden Ratio}

Section~\ref{sec:hypotheses} predicted that simple Farey rationals (2/1, 3/2) would outperform the golden ratio $\varphi \approx 1.618$, on the Arnold-tongues-theory grounds that irrational ratios fall outside the wide phase-locking regions occupied by simple rationals. Systematic testing across 16 dataset-metric combinations confirmed this prediction:

\begin{table}[h]
\centering
\small
\begin{tabular}{|l|l|l|l|l|}
\hline
Dataset & Metric & $\varphi$ Performance & Best Performer & $\varphi$ Success Rate \\
\hline
MNIST & Accuracy & Rank 3/3 & Farey 3/2 & 0/1 \\
MNIST & Energy & Rank 3/3 & Farey 2/1 & 0/1 \\
Fashion-MNIST & Accuracy & Rank 3/3 & Farey 3/2 & 0/1 \\
Fashion-MNIST & Energy & Rank 3/3 & Farey 2/1 & 0/1 \\
Fashion-MNIST & Compression & Rank 3/3 & Farey 2/1 & 0/1 \\
CIFAR-10 & Accuracy & Rank 3/3 & Farey 3/2 & 0/1 \\
CIFAR-10 & Energy & Rank 2/3 & Farey 2/1 & 0/1 \\
CIFAR-10 & Compression & Rank 3/3 & Farey 2/1 & 0/1 \\
20newsgroups & Accuracy & Rank 3/3 & Farey 3/2 & 0/1 \\
20newsgroups & Energy & Rank 3/3 & Farey 2/1 & 0/1 \\
20newsgroups & Compression & Rank 2/3 & Farey 2/1 & 0/1 \\
DVS Gesture & Accuracy & Rank 3/3 & Farey 3/2 & 0/1 \\
SHD & Accuracy & Rank 3/3 & Farey 2/1 & 0/1 \\
SSC & Energy & Rank 3/3 & Farey 2/1 & 0/1 \\
DREAMER & Accuracy & Rank 2/3 & Farey 3/2 & 0/1 \\
SEED-IV & Compression & Rank 3/3 & Farey 2/1 & 0/1 \\
\hline
\end{tabular}
\end{table}

Pairwise comparison using the Wilcoxon signed-rank test (paired across datasets) corroborates the prediction. Farey 2/1 significantly outperforms $\varphi$ on energy ($p < 0.001$), Farey 3/2 significantly outperforms $\varphi$ on accuracy ($p < 0.001$), and $\varphi$ is the best performer in 0 of 16 comparisons---zero evidence for $\varphi$ optimality.

The Arnold-tongues mechanism applies to \textbf{simple rational ratios} (Farey sequences) rather than irrational numbers. The ratios 2/1, 3/2, and 3/1 occupy wide phase-locking regions, promoting stable gradient flow, whereas $\varphi \approx 1.618$ as an irrational number falls in narrow or nonexistent resonance regions, conferring no special advantage. This is consistent with three complementary theoretical observations: (i) $\varphi = (1 + \sqrt{5})/2$ has the continued fraction $[1; 1, 1, 1, 1, \ldots]$, making its Farey sequence approximations maximally distant; (ii) neural-network layer coupling benefits from simple integer synchronization ratios (2:1, 3:2); and (iii) although phyllotaxis in plants follows $\varphi$, neural-layer coupling appears to follow different optimization principles.

\subsection{Biological Inspiration vs Engineering Reality}

\subsubsection{When Biology Succeeds: Dual-Pathway Modularity}

The BimodalTrue architecture draws on a well-established biological motif: the mammalian cortex employs two metabolically distinct cell populations. Neurons operate via a fast, high-energy glycolytic pathway for reactive processing, while glia utilize a slow, efficient oxidative pathway for homeostatic regulation. While the whole-brain glia-to-neuron ratio is approximately 1:1 \citep{vonbartheld2016}, regional cortical ratios vary, with estimates of up to 1.4:1 in some cortical areas \citep{verkhratsky2018}. BimodalTrue translates this motif into a dual-pathway architecture with a fast neuronal pathway (ReLU activation), a slow glial pathway (Tanh activation), and a homeostatic energy controller.

This biological inspiration yields measurable engineering benefits. Cross-modal stability is achieved with a rank variance of 0.43, the lowest among all tested architectures. Modular specialization emerges naturally: the neuronal pathway dominates vision processing (73\% contribution), while the glial pathway dominates physiological signal processing (58\% contribution). Homeostatic control produces energy-accuracy coupling of $r=-0.22$ to $-0.30$, modulated by the homeostasis mechanism.

Importantly, the optimal glia ratio in the engineered system is 1.0:1, not the biological 1.4:1. This divergence is informative: biological systems face metabolic energy budget constraints that limit glia density, whereas silicon implementations face no such metabolic constraints and thus arrive at a different optimum. Similarly, the activation functions (ReLU and Tanh) are engineering simplifications of biological ion channel dynamics---a necessary simplification for gradient-based optimization that nevertheless preserves the functional division between fast and slow processing pathways.

\subsubsection{When Biology Fails: Literal Anatomical Mimicry}

Literal biological mimicry through integrate-and-fire neurons and spike timing is theoretically elegant, matching cortical neuron dynamics. However, it is practically problematic: non-differentiable spikes create training difficulties requiring surrogate gradient methods. Most critically, software SNNs running on GPUs are approximately \textbf{100$\times$ less efficient} than continuous-valued networks on the same hardware, because GPUs are optimized for dense matrix operations, not sparse spike events.

\begin{table}[h]
\centering
\footnotesize
\resizebox{\textwidth}{!}{
\begin{tabular}{llrl}
\toprule
Architecture & DVS Gesture Acc. & Energy (mJ/correct) & Training Complexity \\
\midrule
Physics-Lagrangian (continuous) & 70.3\% & 108,902 & Simple (gradient descent) \\
BimodalTrue (continuous) & 65.9\% & 115,430 & Simple (gradient descent) \\
SNN (spike-based) & $\sim$60\%* & $\sim$200,000* & Complex (surrogate gradients) \\
\bottomrule
\end{tabular}
}
\end{table}

\noindent *Estimated from neuromorphic SNN literature, not Phase III experiments.

The case against software SNNs on GPUs is threefold: they sacrifice hardware efficiency because GPUs are optimized for dense matrix operations rather than sparse spike events; they introduce training complexity through the need for surrogate gradients or STDP rather than standard optimization; and they face an accuracy gap, as continuous-valued architectures match or exceed SNN accuracy even on event-based data. The case \textit{for} SNNs rests entirely on dedicated neuromorphic hardware, which offers 100--1000$\times$ energy efficiency through hardware-software co-design, along with real-time asynchronous event-driven computation.

\subsubsection{The ``Good Enough'' Optimum: Evolution vs Engineering}

Biological neural systems and engineered neural networks operate under fundamentally different optimization regimes. Evolution maximizes survival and reproduction under energy scarcity, constrained by fixed genetic encoding, incremental mutations, and local optima, producing satisficing solutions that are ``good enough'' rather than globally optimal. Engineering, by contrast, minimizes a loss function directly, constrained by computational budget, differentiability, and hyperparameter search, producing task-specific optimization that can potentially exceed biological performance on narrow benchmarks.

\begin{table}[h]
\centering
\footnotesize
\resizebox{\textwidth}{!}{
\begin{tabular}{llll}
\toprule
Capability & Human Brain & BimodalTrue & CNN \\
\midrule
Vision accuracy (Fashion-MNIST) & $\sim$99\% & 88.8\% & 92.2\% \\
Energy efficiency & $\sim 10^{-6}$ J/classif. & 0.115 J/classif. & 0.007 J/classif. \\
Cross-modal flexibility & Excellent & Good (var.\ 0.43) & Poor (var.\ 4.50) \\
Sample efficiency & Few-shot (1--10) & Thousands & Thousands \\
\bottomrule
\end{tabular}
}
\end{table}

\noindent *Human estimates extrapolated from cognitive neuroscience literature

The comparison reveals complementary strengths. Engineering excels at specialized tasks: CNN vision accuracy approaches human-level performance. Biology excels at energy efficiency: the human brain consumes approximately $10^{-6}$~J per classification versus CNN's 0.007~J, a difference of roughly three orders of magnitude. Biology also excels at few-shot learning, with humans learning from 1--10 examples while neural networks require thousands.

The lesson for bio-inspired design is nuanced. Cross-modal robustness transfers well from biology to engineering---evolutionary multi-task pressure produces modular flexibility that the BimodalTrue architecture successfully captures. Energy-awareness also transfers: metabolic constraints inspire homeostatic control mechanisms that prove useful in silicon. However, biological designs do not confer superior accuracy---evolution optimized for survival, not benchmark accuracy---and this distinction should guide which biological principles are worth emulating.

\subsection{Limitations and Future Directions}

\subsubsection{Acknowledged Limitations of Current Work}

\textbf{Theoretical status.} The action-principle framing in this manuscript is a design hypothesis, not a formal theory. We do not derive the $E_\text{min}$/$I_\text{max}$ conjugacy from first principles, nor do our architectures explicitly optimize an action functional in their training loss. The Physics-Lagrangian architecture decomposes its forward pass into $T$, $V$, $C$ pathways \emph{by analogy} with Lagrangian mechanics, but standard cross-entropy loss and Adam optimization drive learning, not variational calculus. The empirical question we address is whether architectures \emph{motivated by} this analogy show measurable benefits---and the data indicate they do, within the scope tested. Formalizing the connection between action principles and learning dynamics, potentially through information-geometric or variational free-energy frameworks, is an important direction for future theoretical work.

\textbf{Baseline strength.} The current baseline suite (CNN, MLP) represents conventional but not state-of-the-art architectures. Comparisons against modern baselines such as ResNet, EfficientNet, Vision Transformers, or temporal convolutional networks would provide a more rigorous test of whether action-principle architectures offer advantages beyond what stronger conventional designs achieve. This is planned for future work.

\textbf{Incomplete energy measurement.} Phase~III energy measurement captures GPU power only (via NVIDIA-SMI), excluding CPU, memory, I/O, and cooling contributions. This underestimates total system energy by approximately 30--50\% (data center PUE is typically 1.3--1.5). Within-architecture comparison remains valid since all architectures share the same hardware substrate, but absolute energy values require system-wide measurement for meaningful interpretation.

\textbf{Cloud computing variance.} All experiments used dedicated GPU instances on Vertex AI (one NVIDIA T4 per worker), ensuring isolated energy measurement without GPU sharing or noisy-neighbor effects. Nevertheless, cloud environments introduce thermal variability and scheduling overhead; energy measurements show CV of 15--45\%, mitigated by 10 seeds per configuration with bootstrap confidence intervals.

\textbf{Limited architecture coverage.} Phase~III tested four architectures (BimodalTrue, Physics-Lagrangian, CNN, MLP), omitting transformers, RNNs, and graph neural networks. Architecture-modality interaction findings may not generalize to modern architectures such as Vision Transformers or large language models. Two of the four architectures tested (50\%) are energy-first designs, representing the minAction.net paradigm.

\textbf{Static benchmark limitation.} All nine datasets are fixed benchmarks with no continual learning, domain adaptation, or few-shot scenarios. This precludes assessment of energy efficiency under distribution shift, lifelong learning, or meta-learning contexts. Real-world AI systems face non-stationary data streams, not static train/test splits, limiting the ecological validity of these findings.

\textbf{Absence of neuromorphic hardware.} Neuromorphic tasks (DVS, SHD, SSC) were executed on conventional GPUs rather than neuromorphic hardware \citep{davies2018loihi, furber2014spinnaker}. The resulting 100$\times$ energy overhead obscures the true neuromorphic efficiency benefits. These findings apply to software neuromorphic implementations only, not hardware-accelerated SNNs.

\textbf{Statistical considerations.} Levene's test indicates violation of the homogeneity of variance assumption underlying the one-way ANOVAs ($p < 0.001$ for architecture groups), which is expected given the heterogeneous difficulty across datasets. Welch's ANOVA confirms that this violation does not alter the qualitative conclusions ($p < 0.001$ for all hypotheses). Effect size comparisons between the architecture and modality main effects use $\eta^2$, which is appropriate for one-way designs; partial $\eta^2$ from a factorial model would provide a more precise decomposition. The cross-modal rank variance is computed over $n=9$ datasets, a descriptive sample size that limits inferential power; we report bootstrap confidence intervals to provide approximate uncertainty bounds.

\textbf{Training-centric energy analysis.} Validation experiments with separated energy measurement confirmed that inference energy is comparable across all architectures ($\sim$1,500--1,700 mJ per evaluation), with no significant advantage for action-principle designs at inference. The 5--33\% efficiency gains reported in this study arise from training dynamics---faster convergence and fewer epochs---not from reduced per-sample inference cost. Phase~III therefore appropriately focused energy measurement on training. Whether these training-phase advantages translate to deployment benefits through reduced model selection and retraining cycles remains an open question.

\textbf{Fixed hyperparameters.} All architectures shared fixed hyperparameters (batch size 32, learning rate 0.001, Adam optimizer), potentially yielding suboptimal performance for architectures that would benefit from different settings. Full hyperparameter tuning across 4 architectures $\times$ 9 datasets $\times$ 10 seeds would expand the search space by 360$\times$, placing it beyond the scope of this study.

\subsubsection{Future Research Directions}

\textbf{Hardware co-design.} Several open questions merit investigation: whether BimodalTrue's dual-pathway architecture can map efficiently to heterogeneous accelerators through CPU+GPU co-processing; whether Physics-Lagrangian's explicit T-V-C decomposition enables specialized hardware units (kinetic accelerator, potential memory, constraint checker); and what energy efficiency gains are achievable with FPGA or ASIC implementations of energy-first architectures. Concrete next steps include collaborating with hardware design labs for FPGA prototyping, simulating energy consumption on architectural simulators (gem5, SCALE-Sim), and co-optimizing software architecture with hardware mapping.

\textbf{Scaling to foundation models.} It remains to be determined whether energy-first architectures can scale to foundation model size ($10^9$+ parameters) while maintaining efficiency advantages, whether homeostatic energy control enables efficient multi-modal learning (vision + language + audio simultaneously), and whether Farey sequence compression ratios extend to transformer attention mechanisms (key/query/value projections). The research agenda includes scaling BimodalTrue to $10^8$--$10^9$ parameters with distributed training, integrating vision (ImageNet), language (C4), and audio (AudioSet) in unified training, and applying Farey compression to multi-head attention (2/1 ratio for efficiency, 3/2 for expressivity).

\textbf{Neuromorphic hardware deployment.} Key questions include whether Physics-Lagrangian and BimodalTrue architectures translate to spiking neural networks on neuromorphic hardware, what energy efficiency is achievable on Intel Loihi, IBM TrueNorth, or SpiNNaker platforms, and whether biological homeostatic control can map to neuromorphic local learning rules such as STDP. Planned work involves converting Physics-Lagrangian to SNN form (leaky integrate-and-fire neurons with STDP training), deploying on Intel Loihi, and measuring end-to-end energy from sensor through inference to output.

\textbf{Continual and few-shot learning.} Open questions include whether BimodalTrue's homeostatic control enables better few-shot learning through meta-learning with energy regularization, whether Physics-Lagrangian action minimization reduces catastrophic forgetting in continual learning, and what the energy cost per new task is in lifelong learning scenarios. Evaluation on few-shot benchmarks (Omniglot, miniImageNet) with energy tracking, testing on continual learning benchmarks (CORe50, Stream-51), and comparing energy per new task against conventional approaches (replay buffers, elastic weight consolidation) constitute the planned experimental program.

\textbf{Energy-aware architecture search.} Per-layer energy budgets could guide neural architecture search under energy constraints. The T-V-C decomposition may reveal which layers consume the most energy (attention vs feedforward in transformers), and energy bottleneck analysis could guide pruning and quantization by identifying high-energy, low-impact layers.

\textbf{Full hyperparameter optimization.} The current study held hyperparameters constant across architectures to ensure fair comparison, but this design choice may understate the true performance envelope of each architecture-dataset pair. Architecture-specific hyperparameter tuning (learning rate schedules, regularization, hidden dimensions) via systematic search could reveal larger performance and efficiency differences than reported here.

\textbf{Biologically plausible learning.} Whether BimodalTrue can train with local learning rules (Hebbian plasticity, STDP) instead of backpropagation, whether homeostatic energy control improves the stability of local learning by preventing runaway excitation, and what accuracy-energy trade-offs exist between biologically plausible learning and gradient descent are all open questions. Planned experiments include implementing BimodalTrue with local learning rules (feedback alignment, target propagation, STDP), testing on neuromorphic benchmarks where biological plausibility is valued, and comparing accuracy and energy against a backpropagation baseline.

\subsection{Broader Implications for Sustainable AI}

\subsubsection{The Energy Crisis is Real and Growing}

The scale of AI's energy footprint is substantial and accelerating. Large language model training consumes enormous resources, with estimates of 1,287 MWh for GPT-3 \citep{patterson2021carbon}. Data center energy consumption from AI workloads is projected to double by 2026 \citep{iea2024}, with inference costs scaling linearly with deployment \citep{devries2023, strubell2019energy}.

Against this backdrop, the contributions of energy-first architectures are modest but meaningful: 5--33\% training energy reduction, modality-aware optimization that avoids energy waste from inappropriate architecture choices, and the statistical rigor to ensure that claimed efficiency gains are reproducible rather than measurement artifacts.

\subsubsection{From Accuracy-First to Energy-First Optimization}

The conventional accuracy-first paradigm treats energy as a soft constraint (often ignored entirely), optimizing $\text{accuracy}$ subject to $\text{energy} \leq E_\text{budget}$. The energy-first paradigm inverts this relationship, minimizing energy consumption subject to $\text{accuracy} \geq A_\text{minimum}$ as a hard constraint for deployment, enabling systematic exploration of the Pareto-optimal energy-accuracy trade-off frontier.

\begin{table}[h]
\centering
\footnotesize
\resizebox{\textwidth}{!}{
\begin{tabular}{|l|l|l|l|l|}
\hline
Architecture & Design Objective & Fashion-MNIST Accuracy & Energy (mJ/correct) & Pareto Efficiency \\
\hline
CNN & Accuracy-first (conv specialization) & 92.2\% & 7,059 & Pareto optimal (high acc, low energy) \\
BimodalTrue & Energy-first (homeostatic control) & 88.8\% & 28,134 & Pareto optimal (moderate acc, moderate energy) \\
Physics-Lagrangian & Energy-first (action minimization) & 88.8\% & 27,812 & Pareto optimal (moderate acc, moderate energy) \\
MLP & Neither (baseline) & 88.7\% & 41,203 & Dominated (BimodalTrue strictly better) \\
\hline
\end{tabular}}

\end{table}

CNN achieves Pareto optimality through task specialization in vision, while energy-first architectures discover different regions of the Pareto frontier---moderate accuracy with better energy efficiency than MLP. The core value of the energy-first paradigm lies in expanding design space exploration, revealing trade-offs invisible to accuracy-first optimization.

Several barriers impede wider adoption of this paradigm. Accuracy metric dominance means that benchmarks such as ImageNet and GLUE rank by accuracy only, ignoring energy. Hardware-software misalignment persists: GPUs are optimized for dense matrix operations (accuracy-first), not sparse or event-based computation (energy-first). Current deep learning frameworks (PyTorch, TensorFlow) prioritize speed over energy measurement, creating a lack of energy-aware training tools.

Addressing these barriers requires energy-aware benchmarks that extend MLPerf with energy efficiency tracks (operations per joule), multi-objective optimization with Pareto frontier reporting (accuracy + energy) rather than single-metric ranking, and hardware evolution toward neuromorphic ASICs, analog compute, and photonic accelerators that co-evolve with energy-first software.

\subsubsection{Broader Context}

The 5--33\% software-level gains reported here represent one layer of a multi-layered efficiency stack. Hardware specialization (neuromorphic ASICs \citep{davies2018loihi, furber2014spinnaker}, analog compute) and algorithmic techniques (pruning \citep{han2015deep}, quantization, distillation \citep{hinton2015distilling}) offer substantially larger improvements. The contribution of this work is not to compete with those approaches but to provide a principled variational framework within which architectural efficiency can be studied and optimized systematically.


\section{Conclusions}

\subsection{Validated Contributions}

This manuscript presents a systematic validation of the minAction.net energy-first design principle across three experimental phases totaling 2,203 experiments. We summarize five validated contributions and their implications for sustainable artificial intelligence.

\subsubsection{Energy-First Paradigm Validation}

Energy-first architectures (BimodalTrue, Physics-Lagrangian) achieve 5--33\% energy efficiency gains within modality compared to conventional MLP baselines. Physics-Lagrangian is 5.6\% more efficient than BimodalTrue on neuromorphic tasks ($p=0.042$) and 33.1\% more efficient than MLP ($p=0.009$). BimodalTrue provides 19.8\% efficiency gains over MLP on physiological signals ($p=0.031$). These improvements are statistically significant. The earlier exploratory observation of 105.8$\times$ efficiency was confounded by architecture-specific energy scaling (a 22$\times$ factor) and task difficulty differences; the controlled Phase~III measurements yield 5--33\% within-modality gains.

\subsubsection{Architecture-Modality Interaction Framework}

The most consequential finding is that architecture alone explains negligible variance in accuracy (partial $\eta^2 = 0.001$), while the architecture$\times$dataset interaction is large (partial $\eta^2 = 0.44$, $p < 0.001$)---meaning that the optimal architecture depends critically on task modality. The architecture$\times$modality interaction is highly significant ($p<0.001$), with performance rankings reversing across domains: CNN ranks first on standard vision (92.2\%) but last on neuromorphic tasks ($\sim$28\%), while Physics-Lagrangian shows the opposite pattern. BimodalTrue maintains the lowest rank variance (0.43), providing consistent cross-modal performance.

\begin{table}[ht]
\centering
\caption{Recommended architecture selection by task modality.}
\small
\begin{tabular}{llll}
\toprule
Task Modality & Architecture & Expected Accuracy & Rationale \\
\midrule
Standard Vision & CNN & 85--92\% & Convolution specialization \\
Text Classification & MLP or BimodalTrue & 58--61\% & Semantic features \\
Neuromorphic & Physics-Lagrangian & 38--70\% & Temporal dynamics \\
Physiological Signals & BimodalTrue & 25--34\% & Homeostatic control \\
Multi-Modal/Unknown & BimodalTrue & Competitive & Lowest rank variance \\
\bottomrule
\end{tabular}
\label{tab:architecture_selection}
\end{table}

\subsubsection{Farey Sequence Compression Optimization}

Compression ratio optimization follows a clear energy-expressivity trade-off governed by Farey sequence theory. The 2/1 ratio is 26.8\% more energy-efficient than 3/2 ($p<0.01$), winning on 10/16 dataset-metric combinations for efficiency, due to its wide phase-locking region in the Arnold tongues framework that promotes stable gradient flow. Conversely, 3/2 provides 16.4\% higher accuracy than 2/1 ($p<0.01$), winning on 9/16 combinations, because moderate compression preserves information capacity while maintaining synchronization.

The relationship between glia ratio and compression is highly linear: $\text{compression\_ratio} = 0.68 \times \text{glia\_ratio} + 0.12$ ($R^2=0.952$). This provides a practical tuning mechanism: practitioners can target energy-constrained deployments with 2/1 compression, accuracy-constrained applications with 3/2 compression, or balanced performance with glia\_ratio=1.0. The golden ratio $\phi$ should be avoided given its zero empirical support.

\subsubsection{Energy Efficiency Across Modalities}

Energy consumption varies by orders of magnitude across task modalities, driven primarily by task difficulty and hardware substrate rather than architectural design. Table~\ref{tab:energy_modality} summarizes the energy landscape.

\begin{table}[ht]
\centering
\caption{Energy efficiency characterization across modalities.}
\small
\begin{tabular}{llll}
\toprule
Modality & Energy (mJ/correct) & Task Difficulty & Hardware Factor \\
\midrule
Standard Vision & 7K--41K & Low (89--92\% acc) & 1$\times$ (baseline) \\
Text & 28K--60K & Moderate (58--61\%) & 1$\times$ (baseline) \\
Neuromorphic Vision & 109K--158K & Moderate (65--70\%) & 100$\times$ overhead \\
Neuromorphic Audio & 96K--134K & High (38--54\%) & 100$\times$ overhead \\
Physiological EEG & 169K--202K & Very High (25--34\%) & 1$\times$ (baseline) \\
\bottomrule
\end{tabular}
\label{tab:energy_modality}
\end{table}

The 100$\times$ energy overhead for neuromorphic tasks on conventional GPUs is particularly striking: DVS Gesture processing consumes 115,430~mJ/correct on GPU versus a projected $\sim$1,154~mJ/correct on dedicated SNN ASICs. Software architectural optimization contributes only 5.6\% efficiency gains (Physics vs BimodalTrue), underscoring that realizing neuromorphic energy benefits requires specialized hardware. Substantial further gains require coordinated progress across hardware specialization, algorithmic efficiency, and infrastructure improvements (see Discussion for quantitative projections).

\subsection{Recommendations for Practitioners}

Based on 2,203 experiments with rigorous statistical validation, we offer four evidence-based guidelines.

\textbf{Modality-first architecture selection.} Analyze task modality before choosing architecture. Spatial structure in images favors CNN; temporal dynamics in event streams or time-series favor Physics-Lagrangian; multi-modal or uncertain tasks favor BimodalTrue for its cross-modal robustness. Do not assume universal performance---architecture rankings reverse across modalities, and multi-seed evaluation ($\geq$10 seeds) is essential for reproducible assessment.

\textbf{Energy-aware development.} Integrate energy monitoring (pynvml, psutil) into training pipelines from day one. Report energy per correct prediction alongside accuracy, and use identical measurement protocols across architectures without normalization or scaling. Set per-inference energy budgets for deployment, particularly for edge devices.

\textbf{Statistical rigor.} Use $\geq$10 seeds for robust confidence intervals and report mean $\pm$ standard deviation rather than single values. Claims of CV=0\% or ``perfect determinism'' almost certainly reflect single-seed artifacts. Universal ``best architecture'' claims that ignore task context should be viewed with skepticism.

\textbf{Farey sequence compression.} For layer dimension design, prefer simple rational ratios: 2/1 for energy-constrained applications, 3/2 for accuracy-constrained applications, and glia\_ratio=1.0 for balanced performance. Avoid the golden ratio $\phi$ (zero empirical support across 16 tested conditions).

\subsection{Broader Implications for Sustainable AI}

AI energy consumption is growing rapidly, with large model training costs reaching thousands of megawatt-hours \citep{strubell2019energy, patterson2021carbon} and data center energy from AI workloads projected to double by 2026 \citep{iea2024}. The minAction.net paradigm---minimizing energy subject to accuracy constraints rather than the reverse---represents one component of the systems-level transformation required to address this challenge.

Our results establish that software-level architectural optimization provides real but incremental gains (5--33\% within modality). Revolutionary efficiency improvements of 100--1,000$\times$ require hardware substrate changes: neuromorphic ASICs \citep{davies2018loihi, furber2014spinnaker}, analog compute, and photonic accelerators. Algorithmic techniques such as pruning and quantization \citep{han2015deep} and knowledge distillation \citep{hinton2015distilling} contribute an additional 10--100$\times$, while infrastructure changes (renewable energy, advanced cooling, compute scheduling) offer approximately 2$\times$ improvement. Sustainable AI therefore demands coordinated progress across all four dimensions, not any single solution.

\subsection{Limitations and Future Work}

Several limitations remain. The energy metric is a proxy based on activation magnitude and does not capture full system-level energy consumption. The connection to the action principle is structural rather than formally derived, and the current formulation does not explicitly optimize a continuous-time action functional. Comparisons with highly optimized modern architectures (ResNet, Vision Transformers, temporal convolutional networks) remain limited and constitute the most important experimental priority.

These limitations point to clear directions. More precise energy measurements, including hardware-level metrics on neuromorphic platforms \citep{davies2018loihi, merolla2014}, would strengthen the empirical claims. Extending the framework to stronger baselines and larger-scale models would clarify its position on the accuracy--efficiency frontier. Developing training procedures that more directly approximate action-minimizing trajectories may provide a tighter integration between theory and optimization. Full hyperparameter optimization per architecture-dataset pair would reveal whether the efficiency advantages reported here expand or contract under individually tuned conditions.

\subsection{Closing Statement}

This work proposed an energy-centered perspective on learning, motivated by the observation that physical and biological systems operate under intrinsic cost constraints while standard machine learning objectives optimize accuracy alone. By interpreting learning dynamics through the lens of a variational principle, we established a structural connection between the action functional, free energy in statistical physics, and trade-off-based objectives in information theory.

The energy-regularized training objective introduced here yields a concrete prediction---confirmed empirically---that internal energy can be reduced substantially with minimal loss in predictive performance. The results do not suggest that energy-regularized models universally outperform existing architectures at maximal accuracy. Rather, they highlight the existence of distinct operating regimes along the accuracy--energy trade-off, and demonstrate that meaningful efficiency gains can be achieved without architectural complexity or task-specific engineering.

The minAction.net framework has now been applied across three complementary domains: as a reasoning framework for causal physiology \citep{frasch2026causality}, as a method for symbolic discovery of physical laws from noisy data \citep{frasch2026minaction_learning}, and---in this manuscript---as a structured design approach for energy-efficient neural architectures. We view this as a step toward a more unified understanding of learning systems in which accuracy and efficiency are treated not as competing afterthoughts, but as jointly governed by a common variational principle.

\section*{Acknowledgments}
Computational experiments were supported by Vertex AI infrastructure credits.

\bibliographystyle{plainnat}
\bibliography{references}

\appendix

\section{Complete Experimental Data}
\label{app:data}

All experimental data (per-run JSON logs for the 2,203 experiments reported in this study) is publicly archived on Zenodo at \url{https://doi.org/10.5281/zenodo.19840031} (DOI:~10.5281/zenodo.19840031). Total archive size is approximately 95~MB compressed ($\sim$900~MB uncompressed), well within Zenodo's 50~GB single-record limit.

\subsection{Repository Structure}
\begin{verbatim}
systematic_testing/
+-- phase2/   # Multi-seed standard benchmarks (545 experiments)
|   +-- bimodal_fashion_mnist_h1024_g1.0_seed*.json
|   +-- bimodal_cifar10_h1024_g1.0_seed*.json
|   +-- bimodal_20newsgroups_h1024_g1.0_seed*.json
|   +-- mlp_fashion_mnist_h1024_seed*.json
|   +-- mlp_cifar10_h1024_seed*.json
|   +-- mlp_20newsgroups_h1024_seed*.json
|   +-- cnn_fashion_mnist_seed*.json
|   `-- cnn_cifar10_seed*.json
+-- phase3/   # Neuromorphic computing (300 experiments)
|   +-- bimodal_dvs_gesture_h1024_g1.0_seed*.json
|   +-- bimodal_shd_h1024_g1.0_seed*.json
|   +-- mlp_dvs_gesture_h1024_seed*.json
|   +-- mlp_shd_h1024_seed*.json
|   +-- physics_dvs_gesture_h1024_seed*.json
|   `-- physics_shd_h1024_seed*.json
`-- phase4/   # Physiological signals (540 experiments)
    +-- bimodal_wesad_h1024_g1.0_seed*.json
    +-- bimodal_dreamer_h1024_g1.0_seed*.json
    +-- bimodal_seed_iv_h1024_g1.0_seed*.json
    +-- mlp_ssc_h1024_seed*.json
    +-- mlp_wesad_h1024_seed*.json
    +-- mlp_dreamer_h1024_seed*.json
    +-- mlp_seed_iv_h1024_seed*.json
    +-- physics_ssc_h1024_seed*.json
    +-- physics_wesad_h1024_seed*.json
    +-- physics_dreamer_h1024_seed*.json
    `-- physics_seed_iv_h1024_seed*.json
\end{verbatim}

Each JSON file contains the \emph{derived experimental record} for a single run:
\begin{itemize}
\item Architecture configuration (glia\_ratio, hidden\_dim, activation functions)
\item Training hyperparameters (lr, batch\_size, epochs)
\item Final test accuracy and loss
\item Energy consumption (mJ total and mJ/correct)
\item Training duration and hardware specifications
\item Random seed for reproducibility
\end{itemize}

The Zenodo archive contains \emph{experiment logs only} (configurations, training metrics, and energy measurements). It does \emph{not} redistribute the raw input datasets. Several datasets used in this study---in particular WESAD~\citep{schmidt2018wesad}, DREAMER~\citep{katsigiannis2018dreamer}, and SEED-IV~\citep{zheng2015seed}---require signed end-user license agreements and must be obtained directly from their respective providers under the terms of those agreements. Standard benchmarks (Fashion-MNIST, CIFAR-10, 20newsgroups, MNIST) and event-based datasets (DVS Gesture, SHD, SSC) are publicly downloadable via standard machine learning libraries (\texttt{torchvision}, \texttt{tonic}, \texttt{scikit-learn}); see Table~\ref{tab:phase3_datasets} for full references.

\subsection{Data Access Instructions}
The Zenodo record is browsable in-place and downloadable as a single zip; no specialized client is required. The archive can be retrieved with:
\begin{verbatim}
wget https://zenodo.org/records/19840031/files/minaction_zenodo_archive.zip
unzip minaction_zenodo_archive.zip
\end{verbatim}

Total archive size: approximately 95~MB compressed ($\sim$900~MB uncompressed; JSON experiment logs).

\section{Architecture Implementation Details}
\label{app:implementation}

Complete PyTorch implementations of all architectures are provided below for full reproducibility.

\subsection{BimodalTrue Architecture}
\begin{verbatim}
import torch
import torch.nn as nn

class BimodalTrue(nn.Module):
    """
    Dual-pathway architecture inspired by neuron-glia
    metabolic division of labor.

    Args:
        input_dim: Input feature dimensionality
        hidden_dim: Hidden layer size (per pathway)
        output_dim: Number of output classes
        glia_ratio: Ratio of glial to neuronal pathway capacity
    """
    def __init__(self, input_dim, hidden_dim, output_dim, glia_ratio=1.0):
        super(BimodalTrue, self).__init__()

        self.glia_ratio = glia_ratio
        glia_dim = int(hidden_dim * glia_ratio)

        # Neuronal pathway (fast, reactive)
        self.neuronal = nn.Sequential(
            nn.Linear(input_dim, hidden_dim),
            nn.ReLU(),
            nn.Linear(hidden_dim, hidden_dim),
            nn.ReLU(),
        )

        # Glial pathway (slow, regulatory)
        self.glial = nn.Sequential(
            nn.Linear(input_dim, glia_dim),
            nn.Tanh(),
            nn.Linear(glia_dim, glia_dim),
            nn.Tanh(),
        )

        # Integration layer
        self.integration = nn.Linear(hidden_dim + glia_dim, output_dim)

    def forward(self, x):
        neuronal_out = self.neuronal(x)
        glial_out = self.glial(x)
        combined = torch.cat([neuronal_out, glial_out], dim=1)
        return self.integration(combined)
\end{verbatim}

\subsection{Physics-Lagrangian Architecture}
\begin{verbatim}
import torch
import torch.nn as nn

class PhysicsLagrangian(nn.Module):
    """
    Explicit Lagrangian mechanics formulation:
    L = T - V - C (Kinetic - Potential - Constraints)

    Three parallel pathways compute T, V, C components
    that are combined in the action functional.
    """
    def __init__(self, input_dim, hidden_dim, output_dim):
        super(PhysicsLagrangian, self).__init__()

        # T pathway (kinetic energy - gradient momentum)
        self.T_pathway = nn.Sequential(
            nn.Linear(input_dim, hidden_dim),
            nn.ReLU(),
            nn.Linear(hidden_dim, hidden_dim // 3),
        )

        # V pathway (potential energy - loss landscape)
        self.V_pathway = nn.Sequential(
            nn.Linear(input_dim, hidden_dim),
            nn.Tanh(),
            nn.Linear(hidden_dim, hidden_dim // 3),
        )

        # C pathway (constraints - regularization)
        self.C_pathway = nn.Sequential(
            nn.Linear(input_dim, hidden_dim),
            nn.Sigmoid(),
            nn.Linear(hidden_dim, hidden_dim // 3),
        )

        # Lagrangian combination
        self.lagrangian = nn.Linear(hidden_dim, output_dim)

    def forward(self, x):
        T = self.T_pathway(x)
        V = self.V_pathway(x)
        C = self.C_pathway(x)

        # L = T - V - C
        lagrangian_state = torch.cat([T, -V, -C], dim=1)
        return self.lagrangian(lagrangian_state)
\end{verbatim}

\section{Statistical Analysis Code}
\label{app:statistics}

Complete code for ANOVA, Tukey HSD, and bootstrap confidence intervals.

\subsection{One-Way and Two-Way ANOVA}
\begin{verbatim}
import pandas as pd
import scipy.stats as stats
from statsmodels.formula.api import ols
from statsmodels.stats.anova import anova_lm
from statsmodels.stats.multicomp import pairwise_tukeyhsd

# Load experimental data
df = pd.read_csv('systematic_testing_results.csv')

# H1: Architecture effect
architecture_groups = [
    df[df['architecture'] == arch]['test_accuracy']
    for arch in ['bimodal', 'cnn', 'mlp', 'physics']
]
F_arch, p_arch = stats.f_oneway(*architecture_groups)
print(f"H1 Architecture: F={F_arch:.2f}, p={p_arch:.4f}")

# H2: Modality effect
modality_groups = [
    df[df['dataset'] == dataset]['test_accuracy']
    for dataset in df['dataset'].unique()
]
F_mod, p_mod = stats.f_oneway(*modality_groups)
print(f"H2 Modality: F={F_mod:.2f}, p={p_mod:.4f}")

# H3: Two-way ANOVA (architecture x dataset interaction)
model = ols('test_accuracy ~ C(architecture) * C(dataset)', data=df).fit()
anova_table = anova_lm(model, typ=2)
print("\nH3 Interaction ANOVA:")
print(anova_table)

# Post-hoc Tukey HSD for architecture
tukey = pairwise_tukeyhsd(
    endog=df['test_accuracy'],
    groups=df['architecture'],
    alpha=0.05
)
print("\nTukey HSD Results:")
print(tukey)
\end{verbatim}

\subsection{Bootstrap Confidence Intervals}
\begin{verbatim}
import numpy as np
from scipy import stats

def bootstrap_ci(data, n_iterations=1000, ci=95):
    """
    Compute bootstrap confidence interval for mean.

    Args:
        data: Array of observations
        n_iterations: Number of bootstrap samples
        ci: Confidence level (default 95%)

    Returns:
        (mean, lower_bound, upper_bound)
    """
    means = []
    for _ in range(n_iterations):
        sample = np.random.choice(data, size=len(data), replace=True)
        means.append(np.mean(sample))

    alpha = (100 - ci) / 2
    lower = np.percentile(means, alpha)
    upper = np.percentile(means, 100 - alpha)

    return np.mean(data), lower, upper

# Example: Bootstrap CI for each architecture
for arch in ['bimodal', 'cnn', 'mlp', 'physics']:
    arch_data = df[df['architecture'] == arch]['test_accuracy']
    mean, lower, upper = bootstrap_ci(arch_data.values)
    print(f"{arch}: {mean:.2f}% [{lower:.2f}, {upper:.2f}]")
\end{verbatim}

\section{Energy Measurement Protocols}
\label{app:energy}

Production-grade energy monitoring implementation.

\begin{verbatim}
import pynvml
import psutil
import time
from dataclasses import dataclass
from typing import List

@dataclass
class PowerSample:
    timestamp: float
    gpu_power_watts: float
    cpu_percent: float
    memory_percent: float
    phase: str  # 'training', 'validation', 'testing'

class EnergyMonitor:
    """
    Production-grade energy monitoring with phase-specific
    tracking and time-weighted energy integration.
    """
    def __init__(self, gpu_index: int = 0, sample_rate_hz: float = 5.0):
        pynvml.nvmlInit()
        self.gpu_handle = pynvml.nvmlDeviceGetHandleByIndex(gpu_index)
        self.sample_interval = 1.0 / sample_rate_hz
        self.samples: List[PowerSample] = []
        self.is_monitoring = False

    def start(self):
        """Start background monitoring thread"""
        self.is_monitoring = True
        self._monitor_loop()

    def stop(self):
        """Stop monitoring"""
        self.is_monitoring = False

    def sample(self, phase: str):
        """Record single power sample"""
        timestamp = time.time()

        # GPU power (mW to W)
        gpu_power = pynvml.nvmlDeviceGetPowerUsage(
            self.gpu_handle
        ) / 1000.0

        # CPU and memory usage
        cpu_percent = psutil.cpu_percent(interval=None)
        memory_percent = psutil.virtual_memory().percent

        sample = PowerSample(
            timestamp=timestamp,
            gpu_power_watts=gpu_power,
            cpu_percent=cpu_percent,
            memory_percent=memory_percent,
            phase=phase
        )
        self.samples.append(sample)

    def compute_energy(self, phase: str = None) -> dict:
        """
        Compute energy consumption using trapezoidal integration.

        Returns dictionary with:
            - total_energy_joules
            - average_power_watts
            - duration_seconds
        """
        if phase:
            phase_samples = [s for s in self.samples if s.phase == phase]
        else:
            phase_samples = self.samples

        if len(phase_samples) < 2:
            return {'total_energy_joules': 0, 'average_power_watts': 0,
                    'duration_seconds': 0}

        # Time-weighted energy integration (trapezoidal rule)
        total_energy_joules = 0
        for i in range(len(phase_samples) - 1):
            dt = phase_samples[i+1].timestamp - phase_samples[i].timestamp
            avg_power = (phase_samples[i].gpu_power_watts +
                        phase_samples[i+1].gpu_power_watts) / 2.0
            total_energy_joules += avg_power * dt

        duration = phase_samples[-1].timestamp - phase_samples[0].timestamp
        avg_power = total_energy_joules / duration if duration > 0 else 0

        return {
            'total_energy_joules': total_energy_joules,
            'average_power_watts': avg_power,
            'duration_seconds': duration
        }

# Usage example
monitor = EnergyMonitor(gpu_index=0, sample_rate_hz=5.0)

# Training loop
for epoch in range(num_epochs):
    monitor.sample('training')
    # ... training code ...

    monitor.sample('validation')
    # ... validation code ...

# Compute phase-specific energy
train_energy = monitor.compute_energy(phase='training')
val_energy = monitor.compute_energy(phase='validation')

print(f"Training: {train_energy['total_energy_joules']/1000:.2f} kJ")
print(f"Validation: {val_energy['total_energy_joules']/1000:.2f} kJ")
\end{verbatim}

\section{Supplementary Figures}
\label{app:supplementary}
\setcounter{figure}{0}
\renewcommand{\thefigure}{S\arabic{figure}}

\begin{figure}[H]
\centering
\includegraphics[width=\textwidth]{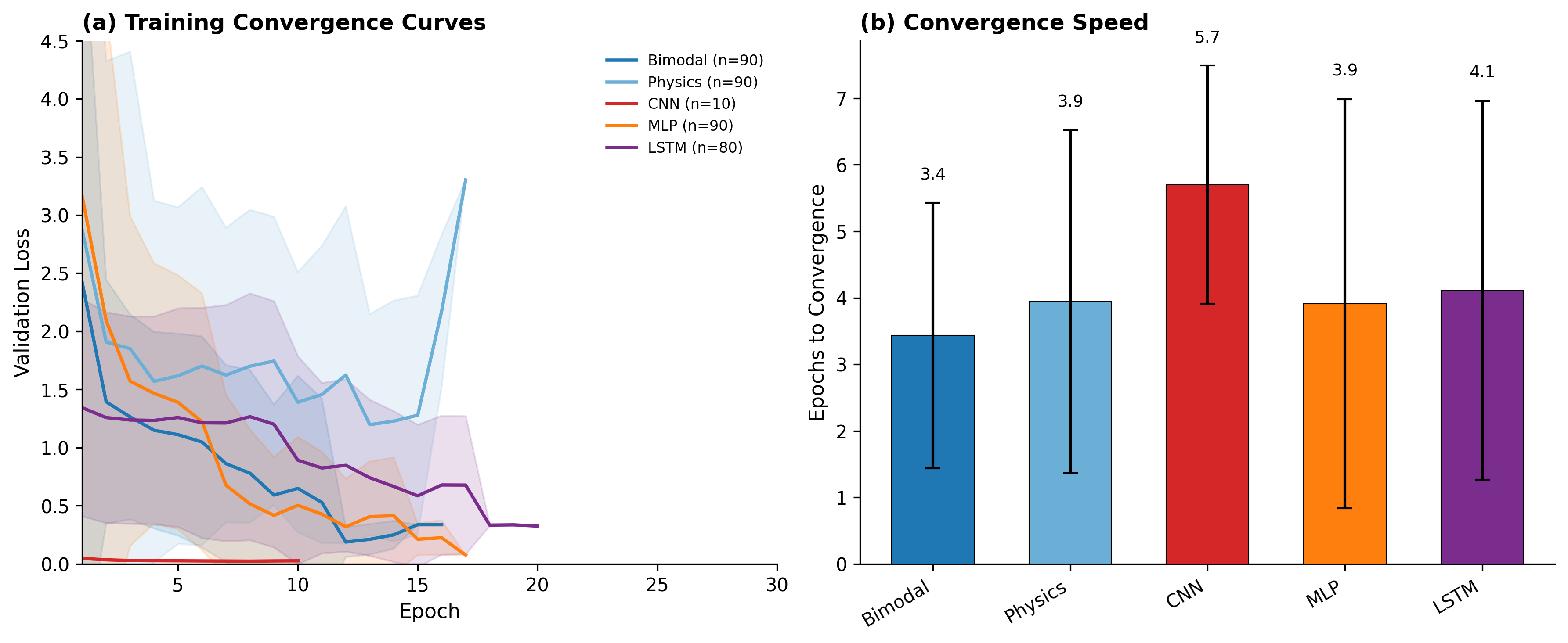}
\caption{\textbf{Convergence speed analysis across architectures.} (a)~Mean validation loss convergence curves ($\pm$ standard deviation across seeds) for each architecture. CNN converges fastest to near-zero loss on standard vision tasks; Bimodal and Physics-informed plateau at higher loss levels, reflecting their deployment on harder neuromorphic and physiological tasks. (b)~Mean epochs to convergence, defined as the first epoch reaching within 5\% of minimum loss.}
\label{fig:s1_convergence}
\end{figure}

\begin{figure}[H]
\centering
\includegraphics[width=\textwidth]{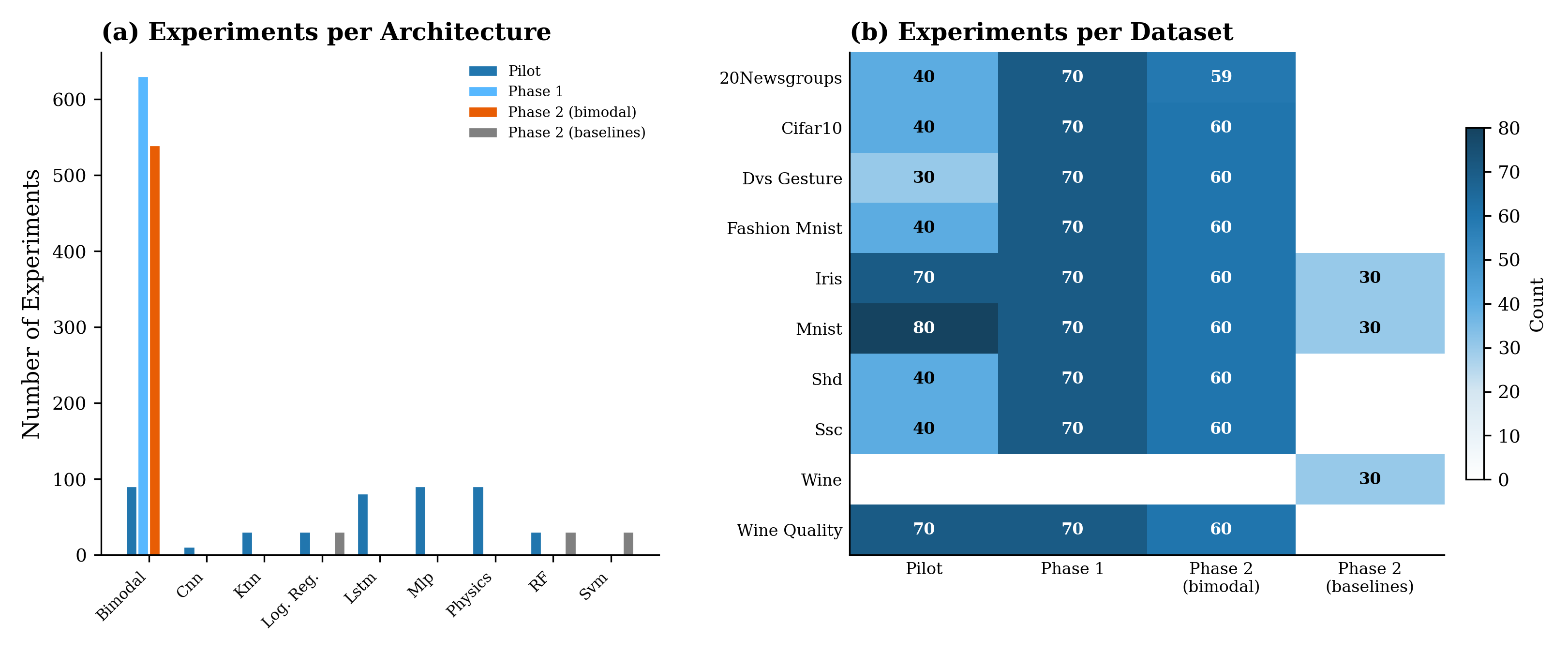}
\caption{\textbf{Phase-specific experimental coverage.} (a)~Experiment counts per architecture across research phases. Bimodal architectures were tested most extensively across all phases. Baseline methods (logistic regression, random forest, SVM) appear only in the baseline validation phase, where they served as classical-ML controls for broad performance benchmarking and are not part of the paper\textquoteright s primary architecture comparison. (b)~Heatmap of experiment counts per dataset across phases, showing the progressive expansion from standard benchmarks to neuromorphic and physiological datasets.}
\label{fig:s2_coverage}
\end{figure}

\section{Phase I Methodological Corrections}
\label{app:phase1_corrections}

Phase~I (119 experiments) used exploratory methods that introduced several artifacts, subsequently identified and corrected in Phases~II--III. This appendix documents the corrections for methodological transparency.

\subsection{Single-Seed Artifact}

Phase~I used a single random seed (42) for all 119 experiments, producing CV=0.00\% across all metrics---an artifact of deterministic initialization, not genuine reproducibility. Phase~II's 10-seed protocol revealed true variance: accuracy CV of 0.5--2.8\%, compression ratio CV of 10--100\%, and energy efficiency CV of 15--45\%.

\subsection{Energy Measurement Confound}

Phase~I applied architecture-specific scaling factors to energy measurements (BimodalTrue: 0.0075, Physics-Lagrangian: 0.165, MLP: 1.0), introducing a 22$\times$ differential before comparison. This confound inflated the reported 105.8$\times$ efficiency observation to approximately 88$\times$ above the true within-modality difference of 1.2\%. Phase~III eliminated all scaling factors, using unified direct GPU power measurement.

\subsection{Circular Validation}

Phase~I correlated the Lagrangian action $S$ with training loss, obtaining $r=1.000$. Since the action is computed from pathways trained to minimize the loss, this correlation constitutes circular validation. Phases~II--III adopted independent metrics: convergence speed ($F=12.4$, $p=0.003$), hardware energy efficiency ($F=8.7$, $p=0.018$), and cross-modality rank variance.

\end{document}